\documentclass{article}

\usepackage{iclr2027_conference,times}
\iclrfinalcopy

\usepackage[utf8]{inputenc}
\usepackage[T1]{fontenc}
\usepackage{hyperref}
\usepackage{url}
\usepackage{booktabs}
\usepackage{amsfonts}
\usepackage{amsmath}
\usepackage{nicefrac}
\usepackage{microtype}
\usepackage{graphicx}
\usepackage{caption}
\usepackage{subcaption}

\usepackage{wrapfig}
\usepackage{needspace}
\usepackage{algorithm}
\usepackage{algpseudocode}
\floatstyle{ruled}
\restylefloat{algorithm}
\newlength{\alglabelwidth}
\algrenewcommand\alglinenumber[1]{\footnotesize\makebox[\alglabelwidth][r]{#1:}}
\newsavebox{\abltablebox}
\usepackage[noabbrev, capitalise]{cleveref}
\IfFileExists{ulem.sty}{\usepackage[normalem]{ulem}}{}

\newcommand{\todo}[1]{}
\usepackage{xcolor}

\definecolor{MyMarkerYellow}{HTML}{FFFACA}

\definecolor{codegreen}{rgb}{0,0.6,0}
\definecolor{codegray}{rgb}{0.5,0.5,0.5}
\definecolor{codepurple}{rgb}{0.58,0,0.82}
\definecolor{backcolour}{rgb}{0.95,0.95,0.92}
\colorlet{reddish}{red!70}

\newcommand{\galst}[1]{}
\newcommand{\ignore}[1]{}
\newcommand{\gal}[1]{}
\newcommand{\eli}[1]{}
\newcommand{\ef}[1]{}

\newcommand{\bd}[1]{}
\newcommand{\ourmethod}{ResizeCO} %
\makeatletter
\def\@maketitle{\vbox{\hsize\textwidth
{\LARGE\sc \@title\par}
\lhead{Preprint}%
\def\And{\end{tabular}\hfill\begin{tabular}[t]{l}\bf\rule{\z@}{24pt}\ignorespaces}%
\hbox to \textwidth{\begin{tabular}[t]{l}\bf\rule{\z@}{24pt}\@author\end{tabular}}%
\vskip 0.3in minus 0.1in}}
\makeatother

\title{Size Generalization for Diffusion Routing Solvers}

\author{Barak Itzchak Davidovitch\textsuperscript{1} \qquad
Eli Meirom\textsuperscript{2} \qquad
Ethan Fetaya\textsuperscript{1,2} \qquad
Gal Chechik\textsuperscript{1,2} \\[2mm]
\textsuperscript{1}Bar-Ilan University \qquad \textsuperscript{2}NVIDIA \qquad \texttt{barakdavidovitch@gmail.com}
}

\begin{document}

\maketitle

\begin{abstract}
\looseness=-1 Routing problems, like the Traveling Salesman Problem (TSP) and the Capacitated Vehicle Routing Problem (CVRP), are NP-hard and slow to solve for large instances. Diffusion-based solvers can be trained in a supervised way using sets of optimal solutions, and then quickly solve new test instances. However, 
 for large instances, optimal solutions are hard to find, and diffusion models fail to generalize to instance sizes outside their training distribution. 
Here we describe \ourmethod{}, a label-free size-generalization approach that addresses both the change in statistical properties of larger instances, and the computational challenge of solving them quickly. It uses reinforcement learning on large-scale instances to adapt a diffusion model trained on small instances. Specifically, we adapt GRPO to this problem, using route length as the training signal. 
\ourmethod{} also formulates $k$-opt as a tree search, solved by a beam search on the GPU. In size-generalization experiments on TSP and CVRP instances with up to 10000 nodes, \ourmethod{} achieves a Pareto front over route quality and time that dominates learned solvers, and also outperforms other solvers at short run times, setting a new state of the art.
\end{abstract}

\section{Introduction}
\label{sec:intro}

\looseness=-1 Combinatorial optimization (CO) problems, and specifically routing problems, appear in many real-world systems, from logistics and supply-chain planning to chip design and scheduling.
Among these, probably the most well-studied and widely deployed are the \textit{Traveling Salesman Problem} (TSP) and the \textit{Capacitated Vehicle Routing Problem} (CVRP), both NP-hard.
For small instances, heuristic approaches can efficiently find near-optimal solutions for these problems, but solving large instances could take hours. Solving within seconds would open new possibilities for real-world problems like on-the-fly routing, interactive “what-if” planning, and large-scale routing, like coordinating deliveries within a city.

\looseness=-1 Supervised learning methods offer a promising approach for fast combinatorial optimization, because the long training cost is amortized over many new instances, each solved quickly at inference \citep{joshi2019efficient, luo2023neural}. Diffusion-based generative solvers, trained by adding noise to precomputed optimal solutions, now match or beat hand-designed heuristics when tested on in-distribution instances \citep{li2025gensco, chen2026maskco}. Unfortunately, supervised training on large instances may be prohibitively expensive because it requires creating a large number of near-optimal solutions as labels. This effect is quantified in \Cref{sec:appendix_label_cost} for TSP and CVRP. At the same time, current supervised approaches also fail at size generalization, namely, diffusion models trained on small instances perform poorly on large instances. Together, scaling supervised training to large instances is challenging.

\looseness=-1 Here we describe \ourmethod{}, a label-free size-generalization approach, designed to generalize from small to large instances. \textbf{\ourmethod{} trains a diffusion model on small instances and then fine-tunes the model using reinforcement learning (RL) on large-scale instances.} %
\ourmethod{} tackles the two main challenges of size generalization: the fact that statistical properties of instances may change with size, and the efficiency challenge of optimizing larger instances. \ourmethod{} addresses these using an integrative approach that handles architecture, fine-tuning and inference.
(1) Architecture: We design a variant of Graph Transformer to generalize better across instance sizes. (2) RL fine-tuning: We adapt GRPO to provide label-free fine-tuning of a diffusion model trained on small instances. Specifically, we use a node-walk decoder, which provides a better estimate of a trajectory probability given the model. (3) Inference: We design a new and efficient \textit{beam $k$-opt search} that runs in parallel on the GPU, which improves inference quality.

\looseness=-1 We evaluate \ourmethod{} on TSP and CVRP and in several size-generalization setups, comparing to both machine-learning and fast non-learned approaches. \ourmethod{} achieves a new state-of-the-art (SoTA) quality for short run times for several size-generalization setups, and dominates learning-based methods over the full Pareto front.

\looseness=-1 Our main contributions are: (1) \ourmethod{} is the first diffusion solver trained on small instances and adapted to large instances without labels, becoming the first diffusion solver trained for CVRP instances of size 10000. (2) Using node-walk decoding, we introduce an exact probability model over the routes sampled from a masked diffusion solver, then used to weight GRPO trajectories. (3) An efficient formulation of $k$-opt as a tree search and a beam-search approach that can be run on the GPU to solve it efficiently. (4) A new SoTA in size generalization on both CVRP and TSP for short durations, and Pareto-domination over learning-based methods in terms of time vs.\ solution quality.

\section{Related Work}
\label{sec:related}

\looseness=-1 \noindent\textbf{Size Generalization in Neural Routing Solvers}
Supervised heatmap solvers, from the graph convolutional network of \citet{joshi2019efficient} to the diffusion-based solvers \citep{sun2023difusco, li2023t2t, li2024fastt2t, li2025gensco, chen2026maskco}, learn from labeled routes. None of them targets size generalization in its main method, so a model trained at one size runs unchanged on larger instances. However, a solver trained on small instances loses accuracy on larger ones \citep{joshi2022learning}. Other methods reach larger sizes with size-robust inductive biases in autoregressive solvers \citep{luo2023neural, drakulic2023bqnco, fang2024invit, huang2025rethinking, gao2024ensemble}, with meta-learning across sizes and distributions \citep{zhou2023omni}, or by splitting a large instance into small subproblems \citep{ye2024glop, hou2023tam, li2021delegate}. In contrast, our method continues training at the target size, without labels.

\looseness=-1 \noindent\textbf{Label-Free Training at the Target Size}
\looseness=-1 Label-free solvers learn from the route length, as a reinforcement learning reward \citep{kool2018attention, kwon2020pomo, qiu2022dimes, greenberg2026reinforcement} or inside an unsupervised loss \citep{min2023unsupervised}, and concurrent work trains autoregressive solvers this way with GRPO \citep{sepulveda2026baseline}. However, these solvers train from scratch, and \citet{joshi2022learning} show that training from scratch on large instances is sample-inefficient, while a model trained on small instances improves further with fine-tuning at the large size. SIT \citep{luo2025boosting} and MaskCO's self-training variant \citep{chen2026maskco} follow this approach, but imitate their own improved routes as pseudo-labels. CADO \citep{song2026cado} fine-tunes a supervised diffusion solver with RL instead, but only at its labeled training sizes. EAS \citep{hottung2022efficient} also fine-tunes a trained model with RL, but per test instance. In contrast, our method fine-tunes a diffusion model supervised on small instances at the target size before test time, with GRPO and the route length as its reward.

\looseness=-1 \noindent\textbf{Size-Generalizing Architectures}
Generative solvers encode instances with a message-passing network on a $k$-nearest-neighbor graph \citep{sun2023difusco, li2024fastt2t} or a transformer \citep{chen2026maskco}, which gives a local or a global view, not both. Graph Transformers keep both by adding graph structure to attention and generalize better than message-passing networks out of distribution \citep{niv2025exploring, wu2025advdifformer}, so we adapt the Graph Transformer of \citet{chen2022structure}.

\looseness=-1 \noindent\textbf{$\boldsymbol{k}$-opt Local Search}
Diffusion-based solvers often post-process (also known as ``polish'') each decoded route with a local operator called 2-opt \citep{croes1958}. It swaps two edges of the route for a shorter pair. LKH-3 \citep{helsgaun2017extension} chains a sequence of edge exchanges into deeper $k$-opt moves with Lin-Kernighan search \citep{lin1973, helsgaun2000effective}, and HGS \citep{vidal2022hybrid} combines 2-opt with other moves. However, a Lin-Kernighan chain grows one exchange at a time. Prior work already parallelized $k$-opt on GPUs, searching all moves of one route in parallel for $k$ up to three \citep{rocki2012accelerating} or six \citep{qiao2026breaking}. In contrast, our method runs Lin-Kernighan chains as a GPU beam search, keeping only the best chains per step for many routes at once.

\section{Preliminaries and Problem Definition}
\label{sec:background}

\looseness=-1 \noindent\textbf{Routing as Edge Selection}
A routing instance is an undirected weighted graph $\mathbb{G}=(V,E)$ with $N=|V|$ nodes and edge lengths $d_{ij}=d_{ji}$. We encode a solution as a flattened binary adjacency matrix $x\in\{0,1\}^{N^2}$, where $x_{ij}=x_{ji}=1$ iff edge $(i,j)$ participates in the route (see also \citealp{sun2023difusco, chen2026maskco}). The length of a route $x$ is $\mathrm{length}(x)=\tfrac{1}{2}d^\top x$. Routing problems minimize this length over $\Omega$, the set of feasible solutions.

\looseness=-1 \noindent\textbf{TSP and CVRP}
In TSP, $\Omega$ holds the routes that visit every node once and return to where they started (Hamiltonian cycles). In CVRP, one node is a depot and every other node $i$ is a customer with a demand $q_i$. A route in $\Omega$ returns to the depot once per vehicle, so it consists of several \textit{sub-routes} from the depot and back. Every customer is on exactly one sub-route, and the total demand on each sub-route is at most the vehicle capacity.

\looseness=-1 \noindent\textbf{Masked Diffusion for CO}
Diffusion-based solvers learn a distribution over solutions by adding noise to optimal solutions during training and predicting that noise.
Our pipeline is based on 
MaskCO \citep{chen2026maskco}. At inference, the denoiser receives a \textit{partial solution} $\hat{x}$ and outputs an \textit{edge heatmap} $p_\theta(\mathbb{G},\hat{x})\in[0,1]^{N^2}$. Starting with all edges masked, a decoder iteratively selects edges from this heatmap over $K$ \textit{unmasking steps}. The decoded route is improved with 2-opt \citep{croes1958}. %
MaskCO can then run multiple correction cycles that re-mask part of the current route, rebuild it, and retain the best solution (see~\cref{fig:method}C). The decoders we use 
are described in~\cref{sec:appendix_inference}.

\looseness=-1 During supervised training, each edge of an optimal route $x^\star$ is masked independently at a rate $p\sim\mathcal{U}(0,1)$, and masking edge $(i,j)$ masks the two entries $x_{ij}$ and $x_{ji}$. For TSP, the loss averages over the $2N$ entries of $x^\star$ that equal one, $
    \mathcal{L}_{\mathrm{SL}}(\theta) = \mathrm{CE}\big(p_\theta(\mathbb{G},\hat{x}), x^{\star}\big) = - \tfrac{1}{2N} \, \smash[b]{\sum_{(i,j) \,:\, x^{\star}_{ij} = 1}} \, \log p_{\theta,ij} .
\looseness=-1 $

\looseness=-1 \noindent\textbf{Group Relative Policy Optimization}
GRPO \citep{shao2024deepseekmath} is a reinforcement learning algorithm that uses a sampled group's mean as its own baseline instead of a value network. For each input, the policy $\pi_\theta$ samples a \textit{group} of $G$ outputs. Output $g$ receives a reward $R_g$, and its \textit{advantage} $A_g$ is the reward standardized within the group,
$
    A_g = \big(R_g - \operatorname{mean}(R_1 \dots R_G)\big) / \operatorname{std}(R_1 \dots R_G). 
$
For generators with $T$ steps, such as flow and masked diffusion models \citep{liu2025flowgrpo, luo2025maskgrpo, zhou2025cogrpo}, GRPO treats generation as a $T$-step Markov decision process with actions $a_t$ taken from states $s_t$ by the policy $\pi_{\theta_{\mathrm{old}}}$ that sampled the group. The loss raises the probability of actions with positive advantage, lowers it for negative ones, and keeps the policy near the \textit{reference} $\pi_{\mathrm{ref}}$, a frozen copy of the starting model:
\begin{equation}
    \begin{aligned}
    \mathcal{L}_{\mathrm{GRPO}}(\theta) &= \frac{1}{G} \sum_{g=1}^{G} \frac{1}{T} \sum_{t=1}^{T} \Big[ - \min\big( r^g_t(\theta)\, A_g , \mathrm{clip}\big(r^g_t, 1{-}\varepsilon, 1{+}\varepsilon\big) A_g \big) + \beta \, D_{\mathrm{KL}}\big( \pi_\theta \,\Vert\, \pi_{\mathrm{ref}} \big)(s_t) \Big] , \\
    r^g_t(\theta) &= \pi_\theta(a_t \mid s_t) \,/\, \pi_{\theta_{\mathrm{old}}}(a_t \mid s_t) .
    \end{aligned}
    \label{eq:grpo_clip}
\end{equation}

\section{Our Method: \ourmethod{}}
\label{sec:method}

\begin{figure}[tb]
\centering
    \includegraphics[width=\linewidth]{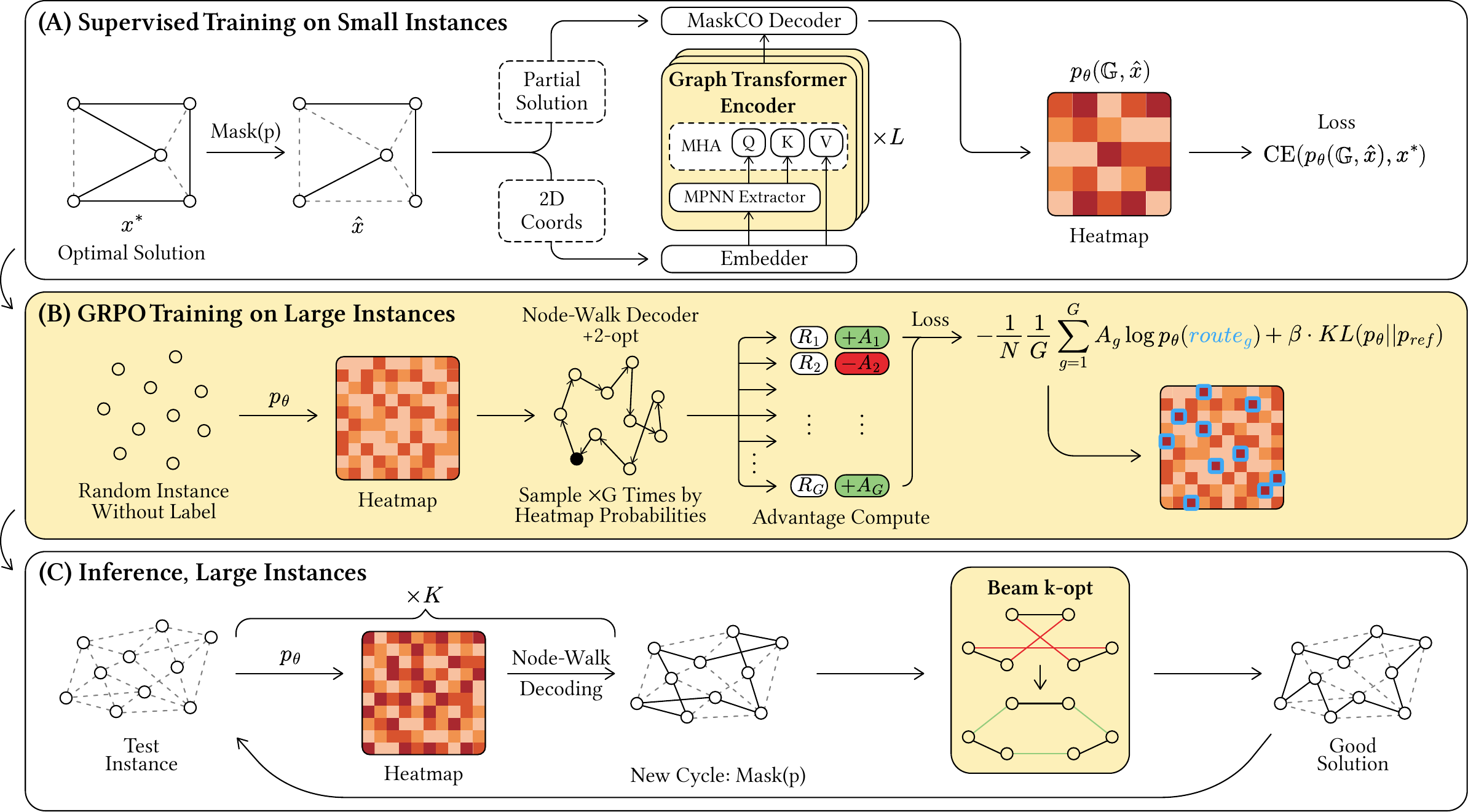}
    \caption{The three stages of \ourmethod{}. 
    \colorbox{MyMarkerYellow}{Yellow marks our novel contributions.} (A) \textbf{Supervised training.} A masked-diffusion model is trained using small labeled instances with a Graph Transformer encoder. (B) \textbf{RL training.} Label-free GRPO training on larger instances. A node-walk decoder samples $G$ routes from the heatmap. Each route is denoted $\mathrm{route}_g$, and their lengths alone determine the advantage.
    (C) \textbf{Inference on large instances.} Multiple correction cycles are executed. Each cycle runs $K$ unmasking steps to build an edge heatmap and decodes the heatmap into a route with the node-walk decoder. The route is then ``polished'' with our GPU beam $k$-opt search. Before the next cycle, the route is partly re-masked.}
    \label{fig:method}
\end{figure}

\looseness=-1 To allow diffusion-based combinatorial optimization to generalize to large instances, we need to address two challenges: 
(1) \textit{Size generalization} -- handling instances that come from a distribution of large instances; and (2) \textit{computational scalability} -- handling larger search spaces. We take a three-pronged approach to handle these. (A) An architecture that supports a scalable representation that captures global structures, while emphasizing local structure that stays meaningful as instances grow (\cref{sec:method_gt}), (B) training for \textit{size generalization without ground-truth labels}, learning from route length alone at the target size (\cref{sec:method_rl}), and (C) \textit{fast parallel inference} whose quality scales with the test-time budget (\cref{sec:method_inference}). \Cref{fig:method} draws the three as one pipeline, with our contributions marked in yellow.

\subsection{The Graph Transformer Encoder: Local Structure in Every Layer}
\label{sec:method_gt}

\looseness=-1 Current approaches \citep{chen2026maskco} encode the routing graphs using transformers, since they capture global structure and are fast. However, transformers can be less sensitive to local structure. This can hurt size generalization significantly, because the distribution of local structures may be preserved when moving to larger instances \citep{fang2024invit}.
We therefore design a variant of \textit{Graph Transformer} for the current optimization problem.

\looseness=-1 Specifically, we construct, for every instance, a $k$-nearest-neighbor graph $\mathbb{G}_k$ from its node coordinates, and assign weights $w_{ij}$ to the edges of that graph, based on the distances between the nodes, $w_{ij} = 1 / (1 + \lVert v_i - v_j \rVert)$ for nodes $i$ and $j$. This weighting assigns larger weight to closer nodes. 

\looseness=-1 We then follow the Graph Transformer architecture of \citet{chen2022structure}. 
We build a two-layer \textit{MPNN extractor} that receives as input the graph $\mathbb{G}_k$, where each layer has the form $h' = h + W_{\mathrm{out}} \cdot\nobreak \mathrm{RMSNorm}\big( \mathrm{SiLU}( W_{\mathrm{self}} \cdot\nobreak h + W_{\mathrm{neigh}} \cdot\nobreak \operatorname{mean}_{j \in \mathcal{N}_k(i)} w_{ij} \cdot\nobreak h_j ) \big)$. This is fully defined in Appendix~\ref{sec:appendix_arch}.
We initialize $W_{\mathrm{out}}$ to zero before supervised training, so the encoder starts as a plain transformer, and train it in both the supervised and the RL stages.
In each attention layer, we then compute the queries and the keys from the extractor output and, as in a standard transformer, the values from the plain node states. This makes the attention weights sensitive to the local structure around each node. All $L$ encoder layers take this form, each running the extractor anew, and the decoder is untouched.

\begin{figure}[!b]
\centering
    \includegraphics[width=0.8\linewidth]{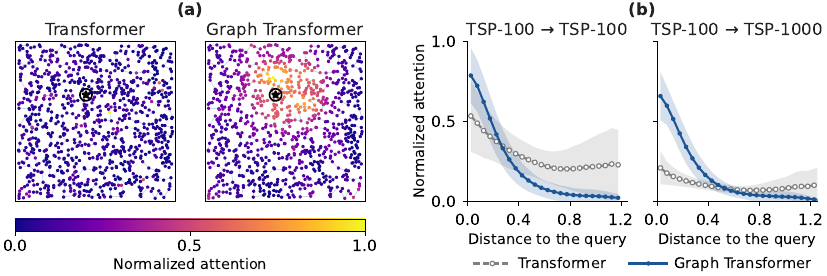}
    \caption{Encoder attention, averaged over all layers and normalized per row. (a) One query node (star) at TSP-1000. (b) Mean attention by distance to the query node. Shaded bands denote std over query nodes.}
    \label{fig:enc_locality}
\end{figure}

\looseness=-1 We analyze the effect of the architecture change on the distribution of encoder attention, comparing 
the transformer and the Graph Transformer, trained on TSP-100, and then tested on TSP-100 (in distribution) and on TSP-1000 (size generalization), see \cref{fig:enc_locality}. Both models attend more strongly to close neighbors when tested in distribution (\cref{fig:enc_locality}b left $100 \rightarrow 100$), but only the Graph Transformer keeps that preference in size generalization (\cref{fig:enc_locality}b right $100 \rightarrow 1000$). This demonstrates how the locality preference learned by the Graph Transformer survives the size change.

\Needspace{25\baselineskip}
\subsection{Label-Free GRPO at the Target Size}
\label{sec:method_rl}

\Needspace{22\baselineskip}
\begin{wrapfigure}{r}{0.5\linewidth}
    \vspace{-\intextsep}
    \begin{minipage}{\linewidth}
    \vspace{-\intextsep}
    \begin{algorithm}[H]
    \captionsetup{font=normalsize}
    \caption{Label-free GRPO at the target size. Blue text marks our changes to standard GRPO.}
    \label{alg:grpo}
    \scriptsize
    \begin{algorithmic}[1]
    \Statex \hspace*{-1.5em}\parbox[t]{\dimexpr\linewidth+1.5em\relax}{\algorithmicrequire\ $\mathcal{D}$ unlabeled instances, $\theta_0$ supervised model, $G$ group size, $K$ unmasking steps per route, $\eta$ \mbox{learning rate}}
    \State $\theta \gets \theta_0$, \ $p_{\mathrm{ref}} \gets p_{\theta_0}$ \Comment{frozen}
    \Repeat
    \State draw an instance $\mathbb{G} \sim \mathcal{D}$ with $N$ nodes
    \For{$g = 1 \dots G$} \Comment{distinct start nodes}
    \State $\hat{x}_g \gets \emptyset$, \ $n_1 \gets$ the $g$-th start node, \ $U_1 \gets V \setminus \{n_1\}$
    \For{$t = 1 \dots K$}
    \State $p^{t}_\theta \gets p_\theta(\mathbb{G}, \hat{x}_g)$ \Comment{one forward pass}
    \For{$j = \lfloor (t-1)N/K \rfloor + 1 \dots \lfloor tN/K \rfloor$}
    \State \textcolor{blue}{draw $n_{j+1} \sim \pi_\theta(\cdot \mid s_j ; p^{t}_\theta)$ \Comment{Eq.~\eqref{eq:walk-step}}}
    \State \textcolor{blue}{$\hat{x}_g \gets \hat{x}_g \cup \{(n_j, n_{j+1})\}$ \Comment{unmask the edge}}
    \State \textcolor{blue}{$U_{j+1} \gets U_j \setminus \{n_{j+1}\}$}
    \EndFor
    \EndFor
    \State $\mathrm{route}_g \gets \hat{x}_g$
    \State \textcolor{blue}{$R_g \gets -\,\mathrm{length}\big(\text{2-opt}(\mathrm{route}_g)\big)$}
    \EndFor
    \State $A_g \gets$ advantage of $R_1 \dots R_G$
    \State $\theta \gets \theta - \eta \nabla \mathcal{L}_{\mathrm{\ourmethod}}(\theta)$ \Comment{one update, Eq.~\eqref{eq:grpo}}
    \Until{the step budget runs out}
    \end{algorithmic}
    \end{algorithm}
    \end{minipage}
    \vspace{-\intextsep}
\end{wrapfigure}

\looseness=-1 Our second main contribution addresses size generalization with no ground-truth labels. True labels are not available for large instances because finding their optimal solutions is costly. We therefore take a reinforcement learning approach that uses the cost of a route as the reward. We adapt GRPO to CO with three changes:
(a) a reward that includes the 2-opt polish operator, (b) a node-walk decoder that defines an exact probability model over routes, and (c) a loss with a single update per batch and a KL penalty on the whole heatmap.

\looseness=-1 First, let us formally define the RL problem, adapting GRPO for masked diffusion to our combinatorial optimization case. We view the construction of one route as a Markov decision process with one decision per edge, and each of the $K$ unmasking steps supplies the heatmap for the next $N/K$ decisions. At decision $j \in \{1, \dots, N\}$, \textbf{the state} $s_j = (\mathbb{G}, \hat{x}_j)$ contains the instance and the edges unmasked so far. The \textbf{action} $a_j \sim \pi_\theta(\cdot \mid s_j ; p^{t}_\theta)$ selects which single edge of the route is unmasked next from the heatmap $p^{t}_\theta = p_\theta(\mathbb{G}, \hat{x}_t)$ of the most recent unmasking step $t$. \Cref{alg:grpo} presents the full training procedure, where the outer loop runs over the unmasking steps $t$ and the inner loop over the decisions $j$.

\looseness=-1 \textbf{(a) The polished cost as a reward.} 
Unlike previous approaches in this area, we use as reward the negative route length \textit{after} a 2-opt polish operator, $R_g = -\,\mathrm{length}\big(\text{2-opt}(\mathrm{route}_g)\big)$.
This provides two benefits: it is more consistent with inference, and 
it removes errors accumulated during iterative node-walk decoding. Here, the advantage ranking corresponds better to solution quality. 

\looseness=-1 \noindent\textbf{(b) Node-walk decoder.} This decoder constructs a route sequentially, node by node. A random node is set as the root. Then, at each step, another node is sampled using the heatmap $p^{t}_\theta$, which represents a conditional distribution because a softmax was applied to each row. The probability of each decision is therefore 
\begin{equation}
    \pi_\theta(a_j \mid s_j ; p^{t}_\theta) = \frac{p^{t}_{\theta,\, n_j n_{j+1}}}{\sum_{n' \in U_j} p^{t}_{\theta,\, n_j n'}},
    \label{eq:walk-step}
\end{equation}
where $n_{j+1}$ is the node at the end of the edge $(n_j, n_{j+1})$ selected at decision $j$, and $U_j$ is the set of nodes not yet visited, except $U_N = \{n_1\}$, so the last decision closes the route with probability one.
Multiplying the decisions within a step and over steps, we obtain the total probability of $\mathrm{route}_g$
\begin{equation}
    \log p_\theta(\mathrm{route}_g) = \sum_{t=1}^{K} \, \sum_{j=\lfloor (t-1)N/K \rfloor + 1}^{\lfloor tN/K \rfloor} \log \pi_\theta(a_j \mid s_j ; p^{t}_\theta) .
    \label{eq:walk}
\end{equation}
In contrast with other masked generative models \citep{luo2025maskgrpo, zhou2025cogrpo}, which score all positions unmasked in the same step as independent, our approach scores masked edges sequentially, each edge being conditioned on its predecessor. As a result, Eq.~\eqref{eq:walk} provides the exact Markovian probability of a route under the node-walk policy, rather than an approximation.

\looseness=-1 Node-walk decoding differs from the greedy decoding that is commonly used in the field. In greedy decoding, all $N^2$ edges of the graph are ranked by their heatmap score, and edges are selected one by one as long as they are feasible. This decoding does not allow us to compute the exact route probability, because it does not compute a proper normalization. GRPO improves the model with node-walk decoding and not with greedy decoding (\cref{sec:method_walk}).

\looseness=-1 \noindent\textbf{(c) Our training procedure and loss.} 
Our loss differs from the standard GRPO loss of Eq.~\eqref{eq:grpo_clip} in two ways (see~\cref{sec:appendix_loss}). First, since rollouts are fast to compute, we take a single update per batch on fresh rollouts.
Second, the KL penalty is the row average over the whole edge heatmap rather than over the visited states only.
This yields our final loss
\begin{equation}
    \mathcal{L}_{\mathrm{\ourmethod}}(\theta) = - \frac{1}{N} \frac{1}{G} \sum_{g=1}^{G} A_g \log p_\theta(\mathrm{route}_g) + \beta \, D_{\mathrm{KL}}\big( p_\theta \,\Vert\, p_{\mathrm{ref}} \big) .
    \label{eq:grpo}
\end{equation}

\subsection{Accelerating Inference Using a GPU Beam \texorpdfstring{$k$}{k}-opt Search}
\label{sec:method_inference}

\looseness=-1 We now describe the third technical contribution, which accelerates inference using a GPU beam $k$-opt search.
At inference, a decoded route usually still contains local mistakes that a few edge exchanges can fix, so every route ends in a local search. MaskCO uses 2-opt for this search, and we replace it with the stronger $k$-opt.

\looseness=-1 The $k$-opt procedure is a highly useful sequential procedure.
Simply put, in each move $k$ edges are removed, and one aims to add back another set of $k$ edges that improves the cost. For example, 2-opt replaces a pair of crossing edges with non-crossing edges. 
While $k$-opt is highly useful, finding sets of $k$ edges that improve the cost can be computationally expensive.

\looseness=-1 Following LKH \citep{helsgaun2000effective}, we simplify and view the $k$-opt process as a series of decisions that form a tree. The root of the tree is the current solution (for TSP, a Hamiltonian cycle). It has $\Theta(N^2)$ outgoing tree branches. Moving from the root to one of these branches means that we remove an edge $(n_i,n_j)$ in the solution (one among $N$ edges of the solution), and add another edge, outgoing from $n_i$ or $n_j$ (say, $n_i$), to another node (the ``front node''). This breaks the validity of the solution. Every deeper level removes one of the two solution edges of the last ``front node'', and adds another edge, from the other node of that removed edge to another node, which becomes the new ``front node'', or back to $n_j$. Looping back to $n_j$ closes the route, which is feasible again if it is a single cycle. $k$-opt means that we look for solutions that are feasible at depth at most $k$. LKH searches the tree in a sequential way, which can be thought of intuitively as a depth-first approach, and continues as long as it finds a better route.

\looseness=-1 This sequential search fits a GPU poorly, because each decision depends on the one before it and the work differs from route to route. \textbf{We instead search over the tree using a parallel beam search.} At each level, we keep the top $M$ branches active, advance the set of $M$ branches deeper until depth $k$, and also close every branch to record the best feasible solutions. All branches of all the routes in a batch perform the same computation at the same moment, so they are computed in parallel on a GPU. At the end, we apply together all the recorded moves that do not overlap on the route, whereas LKH applies one move at a time. We provide the details in~\cref{sec:appendix_beam}.

\section{Experiments}
\label{sec:experiments}

\looseness=-1 We evaluate \ourmethod{} on TSP and CVRP instances of various sizes, against baselines from four families. Every experiment tests size generalization. We take models trained at one instance size and test them on larger ones.

\label{sec:exp_setup}

\looseness=-1 \noindent\textbf{Datasets}
For synthetic data, we follow the standard data generation protocol of \citet{kool2018attention}, for both TSP and CVRP. Each instance is a set of $N$ nodes sampled uniformly in the unit square. CVRP demands are integers drawn uniformly from 1 to 9, with a vehicle capacity of 50 \citep{chen2026maskco}. %
The training and test sets cover sizes 100, 500, 1000 and 10000, and only the supervised stage uses the solutions that come with the training sets. For real-world instances, we test on the benchmark libraries TSPLIB \citep{reinelt1991tsplib} and CVRPLIB \citep{uchoa2017new, arnold2019efficiently}, in three groups per library matched to the test sizes 500, 1000 and 10000. Every set is detailed in~\cref{sec:appendix_datasets}.

\looseness=-1 \noindent\textbf{Metrics}
Following prior work \citep{sun2023difusco, li2025gensco, chen2026maskco}, we report three metrics per test size. (1) \textbf{Objective}: the average total route cost over the test set (a single route for TSP, the sum over all sub-routes for CVRP). (2) \textbf{Gap}: the average relative excess in percent over a near-optimal reference solution, $(\mathrm{cost} - \mathrm{ref})/\mathrm{ref} \times 100$. 
(3) \textbf{Time}: the total wall-clock time to solve all test instances (hardware details in~\cref{sec:appendix_hardware}).

\looseness=-1 \noindent\textbf{Baseline Methods}
We compare \ourmethod{} with four families of solvers.
(1) Classical solvers: an exact solver and the leading heuristic solvers. (2) Supervised learning (SL): autoregressive solvers trained on labeled instances. (3) Label-free learning: solvers trained with RL or self-improvement, without labels. For fair evaluation, they were run with their released models trained nearest to the target size. (4) Diffusion-based solvers, among them MaskCO \citep{chen2026maskco}. MaskCO Self-Train in~\cref{fig:curves_cvrp,fig:curves_tsp} is its self-training variant (\cref{sec:appendix_baselines}). For each baseline, we use the released model and the hyperparameters specified in its original paper, selecting the fastest reported configuration and retaining all refinements and post-processing. Each table row uses the configuration name from the corresponding paper.

\looseness=-1 \noindent\textbf{Evaluation Setting and Hyperparameters}
In the experiments below, the quality of solutions depends on the computational budget $T$, and this in turn depends on three hyperparameters $K, C, S$. $K$ is the number of edge-unmasking steps. The unmasking process is then repeated for $C$ correction cycles, and each correction cycle runs with $S$ parallel samples. Together they yield the total forward-pass budget $T = K \cdot C \cdot S$. For \ourmethod{}, we use $C=1$ and $K=1$ and vary $S$ as the compute budget knob. For MaskCO, we use the optimal values reported by the authors, $S=8$, with $K=1,2$ on TSP and $K=16$ on CVRP, and vary the values of $C$ as the compute budget knob. For all other methods, we use the configurations that the authors provide for the test size.

\subsection{CVRP Results}
\label{sec:exp_cvrp}

\begin{figure}[t]
\centering
\includegraphics[width=0.92\linewidth]{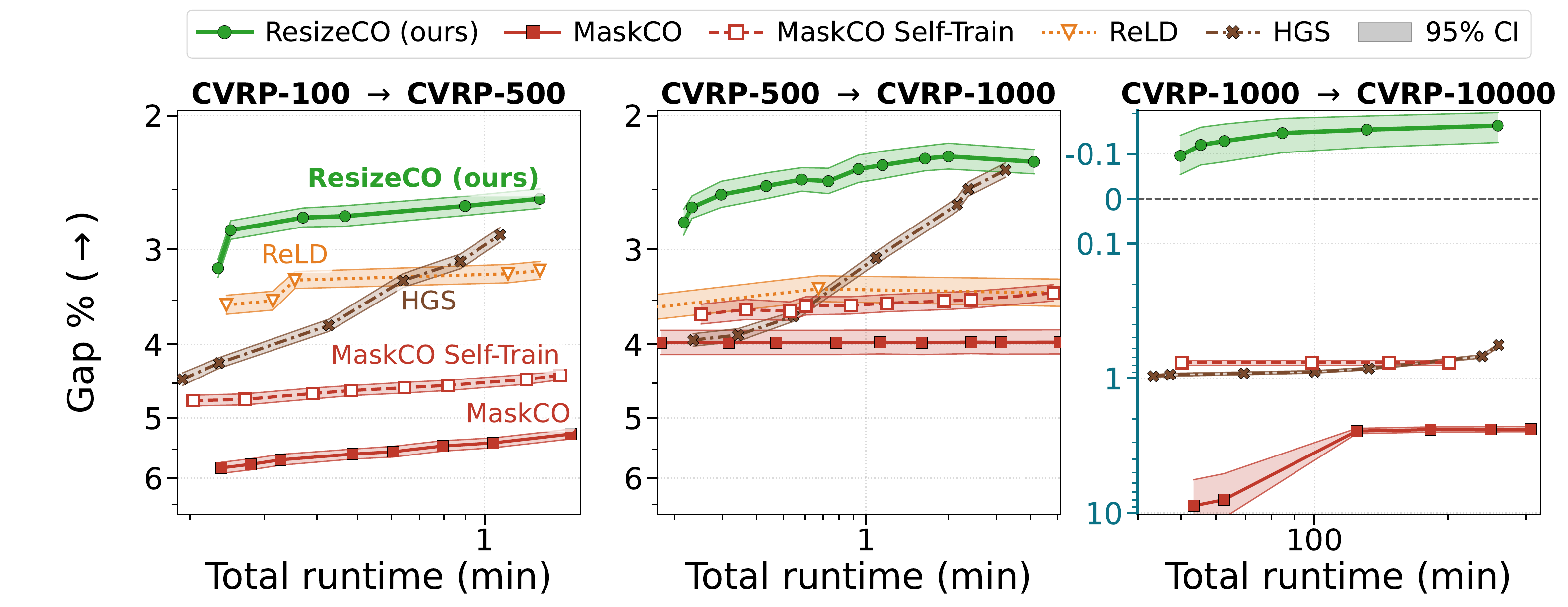}
\caption{CVRP solution quality (gap) against total run time for three source-to-target pairs. Each line varies the compute budget of one method. Our approach \ourmethod{} in green achieves superior quality in short durations. \textbf{From CVRP-1000 to CVRP-10000, \ourmethod{}'s routes are shorter than the HGS reference solutions (negative gap).}}
\label{fig:curves_cvrp}
\end{figure}

\looseness=-1 \Cref{tab:cvrp_main} shows three size-gen\-er\-al\-iza\-tion experiments, each training on a source size and testing on a target size, $100 \rightarrow 500$, $500 \rightarrow 1000$ and $1000 \rightarrow 10000$. In this table, we focus on short inference time, and therefore use configurations reported by competing methods that yielded their fastest results. Among our diffusion-based baselines, only MaskCO supports CVRP, and among the classical solvers we report HGS alone, because LKH-3 stays above 7\% at targets 500 and 1000. \Cref{fig:curves_cvrp} depicts the full Pareto plane of gap against inference run time, focusing on leading approaches.

\begin{table}[!ht]
\caption{CVRP results. Dashes indicate a setting we could not evaluate. An asterisk (*) indicates a number quoted from its paper at a different capacity convention, not comparable to the other rows. $S$ counts parallel samples and $T$ total model forward passes. Bold font marks the best model at a comparable run time.
}
\label{tab:cvrp_main}
\centering
\setlength{\tabcolsep}{3pt}
\resizebox{\linewidth}{!}{%
\begin{tabular}{l|ccc|ccc|ccc}
\toprule
& \multicolumn{3}{c|}{\textbf{CVRP-100 $\rightarrow$ CVRP-500}} & \multicolumn{3}{c|}{\textbf{CVRP-500 $\rightarrow$ CVRP-1000}} & \multicolumn{3}{c}{\textbf{CVRP-1000 $\rightarrow$ CVRP-10000}} \\
\cmidrule(lr){2-4} \cmidrule(lr){5-7} \cmidrule(lr){8-10}
\textbf{Method} & \textbf{Obj} & \textbf{Gap} & \textbf{Time} & \textbf{Obj} & \textbf{Gap} & \textbf{Time} & \textbf{Obj} & \textbf{Gap} & \textbf{Time} \\
\midrule
\multicolumn{10}{l}{\textit{Classical solvers}} \\
HGS (reference) \citep{vidal2022hybrid} & 62.15 & 0.000\% & 55 h & 121.07 & 0.000\% & 197 h & 1120.23 & 0.000\% & 512 h \\
HGS \citep{vidal2022hybrid} & 63.99 & 2.96\% & 52 s & 124.02 & 2.44\% & 2.4 m & 1129.62 & 0.846\% & 2.2 h \\
\midrule
\multicolumn{10}{l}{\textit{Supervised learning}} \\
BQ-NCO (G) \citep{drakulic2023bqnco} & 64.39 & 3.67\% & 26 s & - & - & - & - & - & - \\
LEHD (greedy) \citep{luo2023neural} & 64.40 & 3.63\% & 18 s & - & - & - & - & - & - \\
\midrule
\multicolumn{10}{l}{\textit{Label-free learning}} \\
INViT-3V \citep{fang2024invit} & 67.69 & 9.03\% & 6.1 m & 131.78 & 8.97\% & 10.8 m & 1192.43 & 6.49\% & 2.1 h \\
GLOP-G (LKH-3) \citep{ye2024glop} & - & - & - & - & 8.83\%* & 1.4 m* & - & 8.27\%* & 23.9 m* \\
SIT (PRC10) \citep{luo2025boosting} & - & - & - & - & 4.52\%* & 45 s* & - & 4.43\%* & 14.5 m* \\
ReLD (augx8) \citep{huang2025rethinking} & 64.16 & 3.29\% & 21 s & 125.06 & 3.38\% & 40 s & \multicolumn{3}{c}{OOM} \\
\midrule
\multicolumn{10}{l}{\textit{Diffusion-based}} \\
MaskCO (T=128) \citep{chen2026maskco} & 65.79 & 5.90\% & 11 s & 125.79 & 3.98\% & 11 s & 1197.55 & 7.97\% & 1 h \\
MaskCO (T=1280) \citep{chen2026maskco} & 65.39 & 5.25\% & 1.6 m & 125.79 & 3.98\% & 1.6 m & 1146.56 & 2.36\% & 10.6 h \\
\midrule
\multicolumn{10}{l}{\textit{Ours}} \\
\ourmethod{} (S=1) & \textbf{64.10} & \textbf{3.18\%} & \textbf{14 s} & \textbf{124.35} & \textbf{2.76\%} & \textbf{13 s} & \textbf{1118.83} & \textbf{$-$0.096\%} & \textbf{49.9 m} \\
\bottomrule
\end{tabular}}
\end{table}

\looseness=-1 \textbf{\ourmethod{} reaches the lowest gap of every learned solver in all three experiments, at a similar or lower run time.} Its margin over the strongest learned baseline grows with the target size, and from CVRP-1000 to CVRP-10000 every other learned solver stays far from the reference solutions while \ourmethod{} reaches them, ending below the reference cost as the sample budget grows. Against the classical solver, \ourmethod{} leads at short run times in every experiment (\cref{fig:curves_cvrp}).

\Needspace*{9\baselineskip}
\sbox{\abltablebox}{%
\resizebox{0.618\linewidth}{!}{%
\setlength{\tabcolsep}{3.571pt}%
\begin{tabular}{l cc cc cc}
\toprule
\textbf{CVRPLIB} & \multicolumn{2}{c}{\textbf{100 $\rightarrow$ [321,684]}} & \multicolumn{2}{c}{\textbf{100 $\rightarrow$ [748,956]}} & \multicolumn{2}{c}{\textbf{1000 $\rightarrow$ [7000,15000]}} \\
\cmidrule(lr){2-3} \cmidrule(lr){4-5} \cmidrule(lr){6-7}
& \textbf{Gap} & \textbf{Time} & \textbf{Gap} & \textbf{Time} & \textbf{Gap} & \textbf{Time} \\
\midrule
\ourmethod{} (ours, S=1) & \textbf{3.67\%} & \textbf{0.4 s} & \textbf{3.80\%} & \textbf{0.7 s} & \textbf{8.64\%} & \textbf{30.6 s} \\
MaskCO (T=128) & 6.17\% & 1.1 s & 5.60\% & 1.1 s & 18.37\% & 39 s \\
ReLD (augx8) & 3.76\% & 0.5 s & 4.26\% & 0.9 s & \multicolumn{2}{c}{OOM} \\
\bottomrule
\end{tabular}}}
\begin{wraptable}[10]{r}{\wd\abltablebox}
\captionsetup{width=\wd\abltablebox}
\caption{Mean gap and run time on real-world instances from CVRPLIB. Times are per instance. Bold font marks Pareto-domination (best gap and time) in each column group.}
\label{tab:cvrplib}
\usebox{\abltablebox}
\end{wraptable}
\looseness=-1 \noindent\textbf{Real-World Instances}
We also test on real-world instances from CVRPLIB, whose geometry differs from the synthetic instances. We train on synthetic instances only, so this test changes the geometry as well as the size. \textbf{Averaged over the instances of each group, \ourmethod{} reaches the lowest gap among the learned solvers in all three groups of~\cref{tab:cvrplib}, at a comparable run time.}

\subsection{TSP Results}
\label{sec:exp_tsp}
\begin{figure}[!h]
    \centering
    \includegraphics[width=0.92\linewidth]{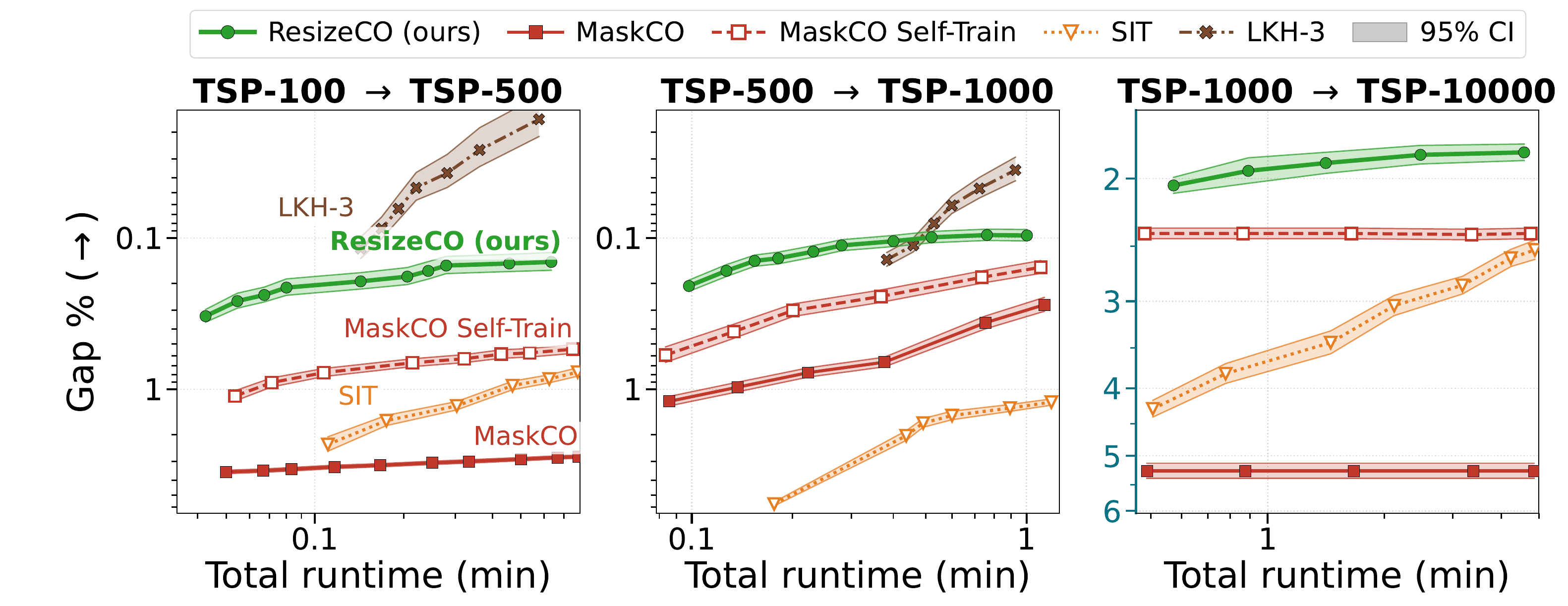}
    \caption{TSP solution quality (gap) against total run time for three source-to-target pairs. Each line varies the compute budget of one method. Our approach \ourmethod{} in green achieves superior quality in short durations. We limit runs on TSP-10000 to less than 5 minutes, which LKH-3 exceeds. Error bars correspond to 95\% confidence intervals.}
    \label{fig:curves_tsp}
\end{figure}

\looseness=-1 \Cref{tab:tsp_xsize} and~\cref{fig:curves_tsp} repeat the three experiments and the short-inference focus of~\cref{sec:exp_cvrp} on TSP. \textbf{At a comparable run time, \ourmethod{} reaches the lowest gap of every learned solver in all three experiments.} From TSP-1000 to TSP-10000, every other learned solver reaches a higher gap, most of them at a far longer run time.

\looseness=-1 Against the classical solver, \ourmethod{} runs faster than LKH-3, and from TSP-500 to TSP-1000 it reaches a lower gap than the fastest LKH-3 setting at a shorter run time (\cref{fig:curves_tsp}).

\begin{table}[!tb]
\caption{TSP size-generalization results. Bold font marks the best model at a comparable run time. ``Time Limit'' denotes cases that did not finish within 4 hours. A dash indicates a setting we did not evaluate. $C$ counts correction cycles and $S$ parallel samples. Inference budgets ($T$, $T_s$, $T_g$) follow each paper's own notation. Larger sample counts lie on the curves of~\cref{fig:curves_tsp}.}
\label{tab:tsp_xsize}
\centering
\setlength{\tabcolsep}{5.714pt}
\resizebox{\linewidth}{!}{%
\begin{tabular}{l|ccc|ccc|ccc}
\toprule
& \multicolumn{3}{c|}{\textbf{TSP-100 $\rightarrow$ TSP-500}} & \multicolumn{3}{c|}{\textbf{TSP-500 $\rightarrow$ TSP-1000}} & \multicolumn{3}{c}{\textbf{TSP-1000 $\rightarrow$ TSP-10000}} \\
\cmidrule(lr){2-4} \cmidrule(lr){5-7} \cmidrule(lr){8-10}
\textbf{Method} & \textbf{Obj} & \textbf{Gap} & \textbf{Time} & \textbf{Obj} & \textbf{Gap} & \textbf{Time} & \textbf{Obj} & \textbf{Gap} & \textbf{Time} \\
\midrule
\multicolumn{10}{l}{\textit{Classical solvers}} \\
Concorde / LKH-3 (reference) & 16.55 & 0.000\% & 37.66 m & 23.12 & 0.000\% & 6.65 h & 71.83 & 0.000\% & 8.8 h \\
LKH-3 (trials=32) \citep{helsgaun2017extension} & 16.55 & 0.026\% & 22 s & 23.13 & 0.036\% & 56 s & 71.81 & $-$0.023\% & 7.7 m \\
LKH-3 (trials=1) \citep{helsgaun2017extension} & 16.57 & 0.118\% & 8.6 s & 23.15 & 0.139\% & 23 s & 71.89 & 0.085\% & 5.6 m \\
\midrule
\multicolumn{10}{l}{\textit{Supervised learning}} \\
BQ-NCO (G) \citep{drakulic2023bqnco} & 16.74 & 1.19\% & 23 s & - & - & - & - & - & - \\
LEHD (greedy) \citep{luo2023neural} & 16.80 & 1.53\% & 15 s & - & - & - & - & - & - \\
\midrule
\multicolumn{10}{l}{\textit{Label-free learning}} \\
INViT-3V \citep{fang2024invit} & 17.27 & 4.38\% & 4.9 m & 24.30 & 5.10\% & 9.8 m & 76.57 & 6.61\% & 33.2 m \\
GLOP \citep{ye2024glop} & 17.07 & 3.20\% & 19 s & 24.01 & 3.85\% & 38 s & 75.61 & 5.27\% & 29 s \\
SIT (PRC10) \citep{luo2025boosting} & 16.70 & 0.947\% & 28 s & 23.39 & 1.18\% & 1.3 m & 74.01 & 3.04\% & 2.1 m \\
\midrule
\multicolumn{10}{l}{\textit{Diffusion-based}} \\
DIFUSCO (T$_s$=50, G) \citep{sun2023difusco} & 17.09 & 3.27\% & 2.2 m & 23.64 & 2.27\% & 9.6 m & 74.11 & 3.18\% & 29.6 m \\
DIFUSCO (T$_s$=100, S=4) \citep{sun2023difusco} & 16.97 & 2.53\% & 21.6 m & 23.48 & 1.56\% & 1.2 h & \multicolumn{3}{c}{Time Limit} \\
Fast T2T (T$_s$=5, T$_g$=5, G) \citep{li2024fastt2t} & 17.30 & 4.58\% & 2.5 m & 23.46 & 1.48\% & 8.9 m & 75.20 & 4.70\% & 3.6 h \\
Fast T2T (T$_s$=5, T$_g$=5, S=4) \citep{li2024fastt2t} & 17.16 & 3.71\% & 11.2 m & 23.37 & 1.10\% & 20.1 m & \multicolumn{3}{c}{OOM} \\
GenSCO (C=1, S=8) \citep{li2025gensco} & 17.19 & 3.88\% & 2.5 s & 23.17 & 0.245\% & 20 s & 89.77 & 24.97\% & 3.4 m \\
GenSCO (C=10, S=8) \citep{li2025gensco} & 17.05 & 3.06\% & 25 s & 23.15 & 0.138\% & 3.3 m & 89.34 & 24.38\% & 34 m \\
MaskCO (T=8/16) \citep{chen2026maskco} & 17.21 & 3.99\% & 1 s & 23.27 & 0.662\% & 23 s & 75.60 & 5.26\% & 3.4 m \\
MaskCO (T=80/160) \citep{chen2026maskco} & 17.07 & 3.18\% & 10 s & 23.15 & 0.138\% & 3.8 m & 75.24 & 4.75\% & 33.9 m \\
\midrule
\multicolumn{10}{l}{\textit{Ours}} \\
\ourmethod{} (S=1) & \textbf{16.60} & \textbf{0.329\%} & \textbf{2.6 s} & \textbf{23.17} & \textbf{0.208\%} & \textbf{5.9 s} & \textbf{73.30} & \textbf{2.05\%} & \textbf{34 s} \\
\bottomrule
\end{tabular}}
\end{table}

\Needspace*{11\baselineskip}
\sbox{\abltablebox}{%
\resizebox{0.643\linewidth}{!}{%
\setlength{\tabcolsep}{3.571pt}%
\begin{tabular}{l cc cc cc}
\toprule
\textbf{TSPLIB} & \multicolumn{2}{c}{\textbf{100 $\rightarrow$ [400,657]}} & \multicolumn{2}{c}{\textbf{100 $\rightarrow$ [724,1432]}} & \multicolumn{2}{c}{\textbf{1000 $\rightarrow$ [11849,14051]}} \\
\cmidrule(lr){2-3} \cmidrule(lr){4-5} \cmidrule(lr){6-7}
& \textbf{Gap} & \textbf{Time} & \textbf{Gap} & \textbf{Time} & \textbf{Gap} & \textbf{Time} \\
\midrule
\ourmethod{} (ours, S=4) & \textbf{1.26\%} & \textbf{0.7 s} & \textbf{1.46\%} & \textbf{1.2 s} & 5.40\% & 20.7 s \\
MaskCO (T=32/32/16) & 3.17\% & 2.1 s & 7.48\% & 1.4 s & 7.23\% & 14 s \\
DIFUSCO (T$_s$=50, G) & 4.62\% & 0.9 s & 5.03\% & 5.3 s & 5.14\% & 4.4 m \\
SIT (PRC10) & 2.24\% & 10.3 s & 2.57\% & 20.4 s & 5.89\% & 27.6 s \\
\bottomrule
\end{tabular}}}
\begin{wraptable}[10]{r}{\wd\abltablebox}
\captionsetup{width=\wd\abltablebox}
\caption{Mean gap and run time on real-world instances from TSPLIB. Times are per instance. Bold font marks Pareto-domination (best gap and time) in each column group.}
\label{tab:tsplib}
\usebox{\abltablebox}
\end{wraptable}
\looseness=-1 \noindent\textbf{Real-World Instances}
\looseness=-1 We also test on real-world instances from TSPLIB. \textbf{Averaged over the instances of each group, \ourmethod{} reaches the lowest gap among the learned solvers in the first two groups of~\cref{tab:tsplib}.} In the last group, DIFUSCO reaches a lower gap at a 13$\times$ longer run time.

\Needspace*{11\baselineskip}
\subsection{Ablation Studies}
\label{sec:exp_ablations}
\sbox{\abltablebox}{%
\resizebox{0.591\linewidth}{!}{%
\setlength{\tabcolsep}{2.857pt}%
\renewcommand{\arraystretch}{0.95}%
\begin{tabular}{l|cc|cc}
\toprule
& \multicolumn{2}{c|}{\textbf{TSP-100 $\rightarrow$ TSP-500}} & \multicolumn{2}{c}{\textbf{TSP-500 $\rightarrow$ TSP-1000}} \\
\cmidrule(lr){2-3} \cmidrule(lr){4-5}
\textbf{Method} & \textbf{Gap} & \textbf{Time} & \textbf{Gap} & \textbf{Time} \\
\midrule
\textbf{\ourmethod{} (full)} & \textbf{0.329\%} & \textbf{2.6 s} & \textbf{0.208\%} & \textbf{5.9 s} \\
\midrule
w/o label-free GRPO & 2.71\% & 5.6 s & 0.774\% & 9.1 s \\
w/o small-model initialization & 2.29\% & 3.6 s & 2.75\% & 10.0 s \\
w/o Graph Transformer encoder & 0.521\% & 1.6 s & 0.269\% & 3.2 s \\
w/o GPU beam $k$-opt search & 0.456\% & 2.5 s & 0.282\% & 6.4 s \\
\bottomrule
\end{tabular}}}

\Needspace*{8\baselineskip}
\begin{wraptable}[7]{r}{\wd\abltablebox}
\vspace{-\intextsep}
\captionsetup{width=\wd\abltablebox}
\caption{Component ablation of \ourmethod{} in size generalization.
Each row removes a single component.}
\label{tab:abl_components}
\usebox{\abltablebox}
\end{wraptable}
\looseness=-1 Finally, we evaluate the contribution of various components of \ourmethod{}. 
\looseness=-1 \Cref{tab:abl_components} shows that removing any component reduces performance (larger gap).  
See more experiments in~\cref{tab:appendix_abl_more}.

\section{Conclusion}
\label{sec:conclusion}

\looseness=-1 We described \ourmethod{}, a diffusion-based approach for routing optimization that generalizes a model trained on small instances to large instances, using RL on large instances without labels. It combines a Graph Transformer encoder, label-free fine-tuning with GRPO and a GPU beam $k$-opt search, and our ablation shows that removing any one of these components increases the gap. In size-generalization experiments on TSP and CVRP instances with up to 10000 nodes, \ourmethod{} reaches the lowest gap of every learned solver at a comparable run time, and on CVRP its margin over the strongest learned baseline grows with the target size. From CVRP-1000 to CVRP-10000, its routes are shorter than the HGS reference solutions. \ourmethod{} also reaches the lowest gap among the learned solvers in five of the six groups of real-world instances from TSPLIB and CVRPLIB, although we train on synthetic instances only. Together, these results set a new SoTA at short run times, which may open new applications like on-the-fly interactive routing.

\subsubsection*{Acknowledgments}
This study was funded by the Israel Science Foundation (ISF grant no.~208075). We thank Amit Edenzon for helping to create \cref{fig:method}.

\bibliographystyle{iclr2027_conference}
\bibliography{references}

\begin{thebibliography}{43}
\providecommand{\natexlab}[1]{#1}
\providecommand{\url}[1]{\texttt{#1}}
\expandafter\ifx\csname urlstyle\endcsname\relax
  \providecommand{\doi}[1]{doi: #1}\else
  \providecommand{\doi}{doi: \begingroup \urlstyle{rm}\Url}\fi

\bibitem[Applegate et~al.(2006)Applegate, Bixby, Chv{\'a}tal, and Cook]{applegate2006traveling}
David~L. Applegate, Robert~E. Bixby, Va{\v s}ek Chv{\'a}tal, and William~J. Cook.
\newblock \emph{The Traveling Salesman Problem: A Computational Study}.
\newblock Princeton University Press, 2006.

\bibitem[Arnold et~al.(2019)Arnold, Gendreau, and S{\"o}rensen]{arnold2019efficiently}
Florian Arnold, Michel Gendreau, and Kenneth S{\"o}rensen.
\newblock Efficiently solving very large-scale routing problems.
\newblock \emph{Computers \& Operations Research}, 107:\penalty0 32--42, 2019.
\newblock ISSN 0305-0548.

\bibitem[Chen et~al.(2022)Chen, O'Bray, and Borgwardt]{chen2022structure}
Dexiong Chen, Leslie O'Bray, and Karsten Borgwardt.
\newblock Structure-aware transformer for graph representation learning.
\newblock In \emph{Proceedings of the 39th International Conference on Machine Learning}, volume 162 of \emph{Proceedings of Machine Learning Research}, pp.\  3469--3489. PMLR, 17--23 Jul 2022.

\bibitem[Chen et~al.(2026)Chen, Li, and Yan]{chen2026maskco}
Lvda Chen, Yang Li, and Junchi Yan.
\newblock Mask{CO}: Masked generation drives effective representation learning and exploiting for combinatorial optimization.
\newblock In \emph{The Fourteenth International Conference on Learning Representations}, 2026.

\bibitem[Croes(1958)]{croes1958}
G.~A. Croes.
\newblock A method for solving traveling-salesman problems.
\newblock \emph{Operations Research}, 6\penalty0 (6):\penalty0 791--812, 1958.

\bibitem[Drakulic et~al.(2023)Drakulic, Michel, Mai, Sors, and Andreoli]{drakulic2023bqnco}
Darko Drakulic, Sofia Michel, Florian Mai, Arnaud Sors, and Jean-Marc Andreoli.
\newblock {BQ-NCO}: Bisimulation quotienting for efficient neural combinatorial optimization.
\newblock In \emph{Advances in Neural Information Processing Systems}, volume~36, pp.\  77416--77429, 2023.

\bibitem[Fang et~al.(2024)Fang, Song, Weng, and Ban]{fang2024invit}
Han Fang, Zhihao Song, Paul Weng, and Yutong Ban.
\newblock {INV}i{T}: A generalizable routing problem solver with invariant nested view transformer.
\newblock In \emph{Proceedings of the 41st International Conference on Machine Learning}, volume 235 of \emph{Proceedings of Machine Learning Research}, pp.\  12973--12992. PMLR, 21--27 Jul 2024.

\bibitem[Gao et~al.(2024)Gao, Shang, Xue, Li, and Qian]{gao2024ensemble}
Chengrui Gao, Haopu Shang, Ke~Xue, Dong Li, and Chao Qian.
\newblock Towards generalizable neural solvers for vehicle routing problems via ensemble with transferrable local policy.
\newblock In \emph{Proceedings of the Thirty-Third International Joint Conference on Artificial Intelligence, {IJCAI-24}}, pp.\  6914--6922. International Joint Conferences on Artificial Intelligence Organization, 2024.
\newblock Main Track.

\bibitem[Greenberg et~al.(2026)Greenberg, Sielski, Linsenmaier, Gandham, Mannor, Fender, Chechik, and Meirom]{greenberg2026reinforcement}
Ido Greenberg, Piotr Sielski, Hugo Linsenmaier, Rajesh Gandham, Shie Mannor, Alex Fender, Gal Chechik, and Eli Meirom.
\newblock Reinforcement learning for initializing genetic algorithms in vehicle routing.
\newblock \emph{Communications {AI} \& Computing}, 1\penalty0 (1):\penalty0 17, 2026.
\newblock ISSN 3091-292X.

\bibitem[Helsgaun(2000)]{helsgaun2000effective}
Keld Helsgaun.
\newblock An effective implementation of the {Lin-Kernighan} traveling salesman heuristic.
\newblock \emph{European Journal of Operational Research}, 126\penalty0 (1):\penalty0 106--130, 2000.
\newblock ISSN 0377-2217.

\bibitem[Helsgaun(2017)]{helsgaun2017extension}
Keld Helsgaun.
\newblock \emph{An Extension of the Lin-Kernighan-Helsgaun TSP Solver for Constrained Traveling Salesman and Vehicle Routing Problems: Technical report}.
\newblock Roskilde Universitet, 2017.

\bibitem[Hottung et~al.(2022)Hottung, Kwon, and Tierney]{hottung2022efficient}
Andr{\'e} Hottung, Yeong-Dae Kwon, and Kevin Tierney.
\newblock Efficient active search for combinatorial optimization problems.
\newblock In \emph{International Conference on Learning Representations}, 2022.

\bibitem[Hou et~al.(2023)Hou, Yang, Su, Wang, and Deng]{hou2023tam}
Qingchun Hou, Jingwei Yang, Yiqiang Su, Xiaoqing Wang, and Yuming Deng.
\newblock Generalize learned heuristics to solve large-scale vehicle routing problems in real-time.
\newblock In \emph{The Eleventh International Conference on Learning Representations}, 2023.

\bibitem[Huang et~al.(2025)Huang, Zhou, Cao, and Xu]{huang2025rethinking}
Ziwei Huang, Jianan Zhou, Zhiguang Cao, and Yixin Xu.
\newblock Rethinking light decoder-based solvers for vehicle routing problems.
\newblock In \emph{The Thirteenth International Conference on Learning Representations}, 2025.

\bibitem[Joshi et~al.(2019)Joshi, Laurent, and Bresson]{joshi2019efficient}
Chaitanya~K. Joshi, Thomas Laurent, and Xavier Bresson.
\newblock An efficient graph convolutional network technique for the travelling salesman problem.
\newblock \emph{arXiv preprint arXiv:1906.01227}, 2019.

\bibitem[Joshi et~al.(2022)Joshi, Cappart, Rousseau, and Laurent]{joshi2022learning}
Chaitanya~K. Joshi, Quentin Cappart, Louis-Martin Rousseau, and Thomas Laurent.
\newblock Learning the travelling salesperson problem requires rethinking generalization.
\newblock \emph{Constraints}, 27\penalty0 (1):\penalty0 70--98, Apr 2022.

\bibitem[Kool et~al.(2019)Kool, van Hoof, and Welling]{kool2018attention}
Wouter Kool, Herke van Hoof, and Max Welling.
\newblock Attention, learn to solve routing problems!
\newblock In \emph{International Conference on Learning Representations}, 2019.

\bibitem[Kwon et~al.(2020)Kwon, Choo, Kim, Yoon, Gwon, and Min]{kwon2020pomo}
Yeong-Dae Kwon, Jinho Choo, Byoungjip Kim, Iljoo Yoon, Youngjune Gwon, and Seungjai Min.
\newblock {POMO}: Policy optimization with multiple optima for reinforcement learning.
\newblock In \emph{Advances in Neural Information Processing Systems}, volume~33, pp.\  21188--21198, 2020.

\bibitem[Li et~al.(2021)Li, Yan, and Wu]{li2021delegate}
Sirui Li, Zhongxia Yan, and Cathy Wu.
\newblock Learning to delegate for large-scale vehicle routing.
\newblock In \emph{Advances in Neural Information Processing Systems}, volume~34, pp.\  26198--26211, 2021.

\bibitem[Li et~al.(2023)Li, Guo, Wang, and Yan]{li2023t2t}
Yang Li, Jinpei Guo, Runzhong Wang, and Junchi Yan.
\newblock {T2T}: From distribution learning in training to gradient search in testing for combinatorial optimization.
\newblock In \emph{Advances in Neural Information Processing Systems}, volume~36, pp.\  50020--50040, 2023.

\bibitem[Li et~al.(2024)Li, Guo, Wang, Zha, and Yan]{li2024fastt2t}
Yang Li, Jinpei Guo, Runzhong Wang, Hongyuan Zha, and Junchi Yan.
\newblock Fast {T2T}: Optimization consistency speeds up diffusion-based training-to-testing solving for combinatorial optimization.
\newblock In \emph{Advances in Neural Information Processing Systems}, volume~37, pp.\  30179--30206, 2024.

\bibitem[Li et~al.(2025)Li, Chen, Wang, Wang, and Yan]{li2025gensco}
Yang Li, Lvda Chen, Haonan Wang, Runzhong Wang, and Junchi Yan.
\newblock Generation as search operator for test-time scaling of diffusion-based combinatorial optimization.
\newblock In \emph{Advances in Neural Information Processing Systems}, volume 38, Main Conference, pp.\  127168--127196, 2025.

\bibitem[Lin \& Kernighan(1973)Lin and Kernighan]{lin1973}
S.~Lin and B.~W. Kernighan.
\newblock An effective heuristic algorithm for the traveling-salesman problem.
\newblock \emph{Operations Research}, 21\penalty0 (2):\penalty0 498--516, 1973.

\bibitem[Liu et~al.(2025)Liu, Liu, Liang, Li, Liu, Wang, Wan, Zhang, and Ouyang]{liu2025flowgrpo}
Jie Liu, Gongye Liu, Jiajun Liang, Yangguang Li, Jiaheng Liu, Xintao Wang, Pengfei Wan, Di~Zhang, and Wanli Ouyang.
\newblock {Flow-GRPO}: Training flow matching models via online {RL}.
\newblock In \emph{Advances in Neural Information Processing Systems}, volume 38, Main Conference, pp.\  40783--40818, 2025.

\bibitem[Luo et~al.(2023)Luo, Lin, Liu, Zhang, and Wang]{luo2023neural}
Fu~Luo, Xi~Lin, Fei Liu, Qingfu Zhang, and Zhenkun Wang.
\newblock Neural combinatorial optimization with heavy decoder: Toward large scale generalization.
\newblock In \emph{Advances in Neural Information Processing Systems}, volume~36, pp.\  8845--8864, 2023.

\bibitem[Luo et~al.(2025{\natexlab{a}})Luo, Lin, Wu, Wang, Tong, Yuan, and Zhang]{luo2025boosting}
Fu~Luo, Xi~Lin, Yaoxin Wu, Zhenkun Wang, Xialiang Tong, Mingxuan Yuan, and Qingfu Zhang.
\newblock Boosting neural combinatorial optimization for large-scale vehicle routing problems.
\newblock In \emph{The Thirteenth International Conference on Learning Representations}, 2025{\natexlab{a}}.

\bibitem[Luo et~al.(2025{\natexlab{b}})Luo, Hu, Fan, Sun, Chen, Xia, Zhang, Chang, and Wang]{luo2025maskgrpo}
Yifu Luo, Xinhao Hu, Keyu Fan, Haoyuan Sun, Zeyu Chen, Bo~Xia, Tiantian Zhang, Yongzhe Chang, and Xueqian Wang.
\newblock Reinforcement learning meets masked generative models: {Mask-GRPO} for text-to-image generation.
\newblock In \emph{Advances in Neural Information Processing Systems}, volume 38, Main Conference, pp.\  108460--108485, 2025{\natexlab{b}}.

\bibitem[Min et~al.(2023)Min, Bai, and Gomes]{min2023unsupervised}
Yimeng Min, Yiwei Bai, and Carla Gomes.
\newblock Unsupervised learning for solving the travelling salesman problem.
\newblock In \emph{Advances in Neural Information Processing Systems}, volume~36, pp.\  47264--47278, 2023.

\bibitem[Niv \& Rabin(2025)Niv and Rabin]{niv2025exploring}
Itay Niv and Neta Rabin.
\newblock Exploring graph-transformer out-of-distribution generalization abilities.
\newblock \emph{arXiv preprint arXiv:2506.20575}, 2025.

\bibitem[Qiao et~al.(2026)Qiao, Cr{\'e}put, Wang, Meng, and Mansouri]{qiao2026breaking}
Wen-Bao Qiao, Jean-Charles Cr{\'e}put, Nan Wang, Kun Meng, and Abdelkhalek Mansouri.
\newblock Breaking {TSP} local search barriers: scalable multi-{GPU} parallelisation of 4-opt, 5-opt, 6-opt and hybrid variable $\lambda$-opt.
\newblock \emph{Expert Systems with Applications}, 296:\penalty0 129110, 2026.
\newblock ISSN 0957-4174.

\bibitem[Qiu et~al.(2022)Qiu, Sun, and Yang]{qiu2022dimes}
Ruizhong Qiu, Zhiqing Sun, and Yiming Yang.
\newblock {DIMES}: A differentiable meta solver for combinatorial optimization problems.
\newblock In \emph{Advances in Neural Information Processing Systems}, volume~35, pp.\  25531--25546, 2022.

\bibitem[Reinelt(1991)]{reinelt1991tsplib}
Gerhard Reinelt.
\newblock {TSPLIB} - {A} traveling salesman problem library.
\newblock \emph{INFORMS J. Comput.}, 3:\penalty0 376--384, 1991.

\bibitem[Rocki \& Suda(2012)Rocki and Suda]{rocki2012accelerating}
Kamil Rocki and Reiji Suda.
\newblock Accelerating 2-opt and 3-opt local search using {GPU} in the travelling salesman problem.
\newblock In \emph{2012 International Conference on High Performance Computing \& Simulation (HPCS)}, pp.\  489--495, 2012.

\bibitem[Sep{\'u}lveda \& Ruz(2026)Sep{\'u}lveda and Ruz]{sepulveda2026baseline}
Carlos~S. Sep{\'u}lveda and Gonzalo~A. Ruz.
\newblock Baseline-free policy optimization for neural combinatorial optimization.
\newblock \emph{arXiv preprint arXiv:2606.10321}, 2026.

\bibitem[Shao et~al.(2024)Shao, Wang, Zhu, Xu, Song, Bi, Zhang, Zhang, Li, Wu, and Guo]{shao2024deepseekmath}
Zhihong Shao, Peiyi Wang, Qihao Zhu, Runxin Xu, Junxiao Song, Xiao Bi, Haowei Zhang, Mingchuan Zhang, Y.~K. Li, Y.~Wu, and Daya Guo.
\newblock {DeepSeekMath}: Pushing the limits of mathematical reasoning in open language models.
\newblock \emph{arXiv preprint arXiv:2402.03300}, 2024.

\bibitem[Song et~al.(2026)Song, Yoon, Lee, Jeong, Lee, and Lim]{song2026cado}
Hyungseok Song, Deunsol Yoon, Kanghoon Lee, Han-Seul Jeong, Soonyoung Lee, and Woohyung Lim.
\newblock {CADO}: From imitation to cost minimization for heatmap-based solvers in combinatorial optimization.
\newblock \emph{Transactions on Machine Learning Research}, 2026.
\newblock ISSN 2835-8856.

\bibitem[Sun \& Yang(2023)Sun and Yang]{sun2023difusco}
Zhiqing Sun and Yiming Yang.
\newblock {DIFUSCO}: Graph-based diffusion solvers for combinatorial optimization.
\newblock In \emph{Advances in Neural Information Processing Systems}, volume~36, pp.\  3706--3731, 2023.

\bibitem[Uchoa et~al.(2017)Uchoa, Pecin, Pessoa, Poggi, Vidal, and Subramanian]{uchoa2017new}
Eduardo Uchoa, Diego Pecin, Artur Pessoa, Marcus Poggi, Thibaut Vidal, and Anand Subramanian.
\newblock New benchmark instances for the {Capacitated Vehicle Routing Problem}.
\newblock \emph{European Journal of Operational Research}, 257\penalty0 (3):\penalty0 845--858, 2017.
\newblock ISSN 0377-2217.

\bibitem[Vidal(2022)]{vidal2022hybrid}
Thibaut Vidal.
\newblock Hybrid genetic search for the {CVRP}: Open-source implementation and {SWAP*} neighborhood.
\newblock \emph{Computers \& Operations Research}, 140:\penalty0 105643, 2022.

\bibitem[Wu et~al.(2025)Wu, Yang, Zeng, and Bronstein]{wu2025advdifformer}
Qitian Wu, Chenxiao Yang, Kaipeng Zeng, and Michael~M. Bronstein.
\newblock Supercharging graph transformers with advective diffusion.
\newblock In \emph{Proceedings of the 42nd International Conference on Machine Learning}, volume 267 of \emph{Proceedings of Machine Learning Research}, pp.\  67959--67985. PMLR, 13--19 Jul 2025.

\bibitem[Ye et~al.(2024)Ye, Wang, Liang, Cao, Li, and Li]{ye2024glop}
Haoran Ye, Jiarui Wang, Helan Liang, Zhiguang Cao, Yong Li, and Fanzhang Li.
\newblock {GLOP}: Learning global partition and local construction for solving large-scale routing problems in real-time.
\newblock In \emph{Proceedings of the AAAI Conference on Artificial Intelligence}, volume~38, pp.\  20284--20292, 2024.

\bibitem[Zhou et~al.(2023)Zhou, Wu, Song, Cao, and Zhang]{zhou2023omni}
Jianan Zhou, Yaoxin Wu, Wen Song, Zhiguang Cao, and Jie Zhang.
\newblock Towards omni-generalizable neural methods for vehicle routing problems.
\newblock In \emph{Proceedings of the 40th International Conference on Machine Learning}, volume 202 of \emph{Proceedings of Machine Learning Research}, pp.\  42769--42789. PMLR, 23--29 Jul 2023.

\bibitem[Zhou et~al.(2025)Zhou, Ni, Chen, Liu, Yue, Wang, Wang, Liu, and Huang]{zhou2025cogrpo}
Renping Zhou, Zanlin Ni, Tianyi Chen, Zeyu Liu, Yang Yue, Yulin Wang, Yuxuan Wang, Jingshu Liu, and Gao Huang.
\newblock {Co-GRPO}: Co-optimized group relative policy optimization for masked diffusion model.
\newblock \emph{arXiv preprint arXiv:2512.22288}, 2025.

\end{thebibliography}

\clearpage
\appendix
\raggedbottom
\crefalias{section}{appendix}
\crefalias{subsection}{appendix}
\numberwithin{table}{section}
\numberwithin{figure}{section}

\section*{Appendix}

This appendix details the datasets (\cref{sec:appendix_datasets}), the implementation of \ourmethod{} (\cref{sec:appendix_params}), the configurations of the correction-cycle baselines (\cref{sec:appendix_baselines}), additional results and ablations (\cref{sec:appendix_results}), and the derivation of the training loss (\cref{sec:appendix_loss}).

\section{Datasets}
\label{sec:appendix_datasets}

\noindent\textbf{Synthetic Instances}
The labeled training sets cover sizes 100, 500 and 1000 for both problems, and the supervised stage uses them with their solutions. The RL stage reuses the same instances without their solutions, and adds instance sets that we generate at size 10000. The labeled training sets and the test sets are the ones released with prior work \citep{sun2023difusco, li2024fastt2t, chen2026maskco}, except the CVRP-10000 test set, which we generate from the same distribution. The test sets cover sizes 100, 500, 1000 and 10000, and every test instance comes with a reference solution, against which we measure the gap. For TSP, the test references are optimal routes from Concorde \citep{applegate2006traveling}, except at TSP-10000, where exact solving is out of reach and the test references come from long LKH-3 runs \citep{helsgaun2017extension}. For CVRP, the test references come from the strong heuristic solver HGS \citep{vidal2022hybrid}, with four-hour runs at CVRP-10000, so a CVRP gap is not a distance from the optimum. \Cref{tab:appendix_datasets} lists the set sizes and the solver behind every test reference and every training label.

\begin{table}[ht]
\caption{Test and training sets. A dash marks a training set without labels.}
\label{tab:appendix_datasets}
\centering
\small
\begin{tabular}{l c cc cc}
\toprule
& & \multicolumn{2}{c}{\textbf{Test}} & \multicolumn{2}{c}{\textbf{Training}} \\
\cmidrule(lr){3-4} \cmidrule(lr){5-6}
\textbf{Problem} & \textbf{Size} & \textbf{Instances} & \textbf{Reference} & \textbf{Instances} & \textbf{Labels} \\
\midrule
TSP & 100 & 1280 & Concorde & 1.26 M & LKH-3 \\
TSP & 500 & 128 & Concorde & 128 k & LKH-3 \\
TSP & 1000 & 128 & Concorde & 64 k & LKH-3 \\
TSP & 10000 & 16 & LKH-3 & 6.4 k & - \\
\midrule
CVRP & 100 & 1280 & HGS & 1.54 M & HGS \\
CVRP & 500 & 128 & HGS & 200 k & HGS \\
CVRP & 1000 & 64 & HGS & 100 k & HGS \\
CVRP & 10000 & 128 & HGS & 20 k & - \\
\bottomrule
\end{tabular}
\end{table}

\noindent\textbf{Real-World Instances}
The benchmark libraries TSPLIB \citep{reinelt1991tsplib} and CVRPLIB \citep{uchoa2017new, arnold2019efficiently} provide real-world instances, and we test on them at the end of~\cref{sec:exp_cvrp,sec:exp_tsp}. Each library contributes three groups matched to the test sizes 500, 1000 and 10000, and a group holds every instance within a factor of 0.7 to 1.5 of its size, 24 TSP and 22 CVRP instances in total. The two smaller CVRPLIB groups also stay within 7 to 14 customers per sub-route, near the 10 of our training convention, and the CVRPLIB 10000 group takes its whole band, with 21 to 100 customers per sub-route. The solvers keep their main-table models, and the budget of each row is given in~\cref{tab:cvrplib,tab:tsplib}. The two smaller groups of each library run the models trained at size 100, and the 10000 groups run the source-1000 models, without further training, as in the main-table experiment from size 1000 to size 10000.

\section{Implementation Details}
\label{sec:appendix_params}

\subsection{Architecture}
\label{sec:appendix_arch}
The structure extractor of \ourmethod{} is a small graph neural network that sits inside every encoder layer of the denoiser. It is a two-layer message-passing neural network (MPNN), where each node updates its state from the states of the neighboring nodes. It runs on the $k$-nearest-neighbor graph $\mathbb{G}_k$, whose edge between nodes $i$ and $j$ carries the weight $w_{ij} = 1 / (1 + \lVert v_i - v_j \rVert)$. One extractor layer updates the node states $h$ as
\begin{equation}
    h' = h + W_{\mathrm{out}} \cdot \mathrm{RMSNorm}\big( \mathrm{SiLU}( W_{\mathrm{self}} \cdot h + W_{\mathrm{neigh}} \cdot \operatorname{mean}_{j \in \mathcal{N}_k(i)} w_{ij} \cdot h_j ) \big) ,
    \label{eq:mpnn}
\end{equation}
with $W_{\mathrm{out}}$ initialized to zero before supervised training, so the extractor starts as the identity and the encoder starts as a plain transformer. At the first encoder layer, $h$ is a learned linear embedding of the input features of each node (\cref{tab:appendix_arch}), and at every later layer it is the output of the layer before. In Eq.~\eqref{eq:mpnn}, $W_{\mathrm{self}}$, $W_{\mathrm{neigh}}$ and $W_{\mathrm{out}}$ are learned weight matrices and $\mathcal{N}_k(i)$ is the set of the $k$ nearest neighbors of node $i$. SiLU is the activation function $\mathrm{SiLU}(x) = x \cdot \sigma(x)$, where $\sigma$ is the logistic sigmoid. RMSNorm divides a vector by the root mean square of all entries and multiplies each entry by a learned gain. An encoder layer runs the extractor of Eq.~\eqref{eq:mpnn} on the node states $h$ and computes
\begin{equation}
    \begin{aligned}
    s &= \mathrm{LN}\big( \mathrm{MPNN}(h,\, \mathbb{G}_k) \big) , \qquad Q = W_{Q} \cdot s , \quad K = W_{K} \cdot s , \quad V = W_{V} \cdot \mathrm{LN}(h) , \\[2pt]
    h' &= h + W_{O} \cdot \mathrm{softmax}\!\left( \frac{Q K^{\top}}{\sqrt{d}} \right) \cdot V , \qquad h'' = h' + \mathrm{FFN}\big( \mathrm{LN}(h') \big) .
    \end{aligned}
    \label{eq:qk}
\end{equation}
All $L$ encoder layers take this form, each running the extractor anew, and the decoder is untouched. The neighborhood size $k$ is the only architectural setting that changes with the test size. \Cref{tab:appendix_arch} lists the parameters.

\begin{table}[ht]
\caption{Architecture parameters.}
\label{tab:appendix_arch}
\centering
\small
\begin{tabular}{l cc}
\toprule
\textbf{Parameter} & \textbf{TSP} & \textbf{CVRP} \\
\midrule
Input features & $2$-D coordinates & $2$-D coordinates $+$ demand \\
Encoder layers & 16 & 16 \\
Decoder layers & 6 & 6 \\
Embedding dimension & 256 & 512 \\
Attention heads & 8 & 8 \\
Structure-extractor GNN layers & 2 & 2 \\
Neighborhood size $k$ & $\max(20,\, N/10)$ & $\max(20,\, N/10)$ \\
\bottomrule
\end{tabular}
\end{table}

\subsection{Supervised Training}
\label{sec:appendix_sl}
The supervised stage trains the denoiser to complete partial solutions on the labeled training sets of~\cref{tab:appendix_datasets}, with the configuration in~\cref{tab:appendix_sl}. A warm start from the previous size means that the TSP-500 model starts from the trained TSP-100 model, and the TSP-1000 model from the trained TSP-500 model.

\begin{table}[ht]
\caption{Supervised training parameters. Values in parentheses are training sizes.}
\label{tab:appendix_sl}
\centering
\small
\begin{tabular}{l cc}
\toprule
\textbf{Parameter} & \textbf{TSP} & \textbf{CVRP} \\
\midrule
Batch size & 256 (100), 128 (500), 16 (1000) & 512 (100), 64 (500), 16 (1000) \\
Training steps & 500000 & 225000 (100), 150000 (500, 1000) \\
Warm start & from scratch (100), previous (500, 1000) & from scratch \\
\bottomrule
\end{tabular}
\end{table}

\subsection{Label-Free GRPO Training}
\label{sec:appendix_rl}
\Cref{tab:appendix_rl} lists the configuration of the RL stage. The batch size and the group size $G$ are the largest that fit in GPU memory at each size. The 2-opt pass inside the reward needs enough steps to remove the sampling error of the node-walk decoder (\cref{sec:method_rl}). It must also stop before the routes of a group converge to the same local optimum, because the advantage then loses its signal. Because a 2-opt step costs more on larger instances, the CVRP runs to size 10000 use fewer steps, and the TSP runs to size 10000 keep the step count but restrict the 2-opt pass to the 20 nearest neighbors of each node.

\begin{table}[ht]
\caption{RL fine-tuning parameters. Values in parentheses are the target size or the source and target sizes of the run.}
\label{tab:appendix_rl}
\centering
\small
\setlength{\tabcolsep}{4pt}
\resizebox{\linewidth}{!}{%
\begin{tabular}{l cc}
\toprule
\textbf{Parameter} & \textbf{TSP} & \textbf{CVRP} \\
\midrule
Group size $G$ & 8, 16 (500 to 1000), 64 (to 10000) & 8, 64 (to 10000) \\
Unmasking steps $K$ & 1 & 1 \\
Sampling temperature & 1.0 & 1.0 \\
KL coefficient $\beta$ & 0.005, 0 (to 10000) & 0.005, 0 (to 10000) \\
Batch size & 64 (500), 24 (1000), 96 (500 to 1000), 4 (to 10000) & 16 (500), 4 (1000), 1 (to 10000) \\
Training steps & 200000, 150000 (500 to 1000), 30000 (to 10000) & 200000, 30000 (to 10000) \\
Reward 2-opt steps & 100 & 100, 10 (to 10000) \\
Reward 2-opt neighbors & all, 20 (to 10000) & all \\
Reward & route length after 2-opt & route cost $+\, \rho \cdot \mathrm{overload}$ after 2-opt \\
Overload penalty $\rho$ & - & 3.0 \\
\bottomrule
\end{tabular}%
}
\end{table}

\subsection{Inference}
\label{sec:appendix_inference}
\noindent\textbf{Decoding the Heatmap}
A decoder turns the edge heatmap into a route. \ourmethod{} uses the \textit{node-walk decoder}, which grows the route one node at a time by sampling the next node from the current node's row of $p_\theta$, restricted to the unvisited nodes. On CVRP, it also respects capacity, skipping customers whose demand exceeds the load left in the vehicle and returning to the depot to open a new sub-route. MaskCO, like most heatmap solvers, uses the \textit{greedy decoder}, which adds edges in decreasing heatmap score, skipping any edge that would give a node a third edge or close the loop early. On CVRP, the greedy decoder applies the two-edge rule to customers only, because the depot takes two edges per sub-route. Either route can then be shortened by \textit{2-opt} local search \citep{croes1958}, which deletes two edges and reconnects the two paths the other way around when that is shorter.

\noindent\textbf{Inference Hyperparameters}
\Cref{tab:appendix_resizeco_cfg} lists the inference settings of every \ourmethod{} row. The decoding configuration is set by the test size, not the source size. At inference, \ourmethod{} launches $S$ runs in parallel, each building one route with the node-walk decoder from its own start node, and returns the shortest of the $S$ routes. Each of the $S$ parallel runs makes one forward pass of the denoiser, that is, $K=1$ unmasking step and $C=1$ correction cycle. The local search that follows decoding, the GPU beam $k$-opt search, is described with its settings in~\cref{sec:appendix_beam}.

\begin{table}[ht]
\caption{\ourmethod{} inference parameters. Values in parentheses are test sizes.}
\label{tab:appendix_resizeco_cfg}
\centering
\small
\begin{tabular}{l cc}
\toprule
\textbf{Parameter} & \textbf{TSP} & \textbf{CVRP} \\
\midrule
Decoder & node-walk & node-walk \\
Unmasking steps $K$ & 1 & 1 \\
Correction cycles $C$ & 1 & 1 \\
Batch size & 16 (500, 1000), 1 (10000) & 16 (500), 8 (1000), 1 (10000) \\
\bottomrule
\end{tabular}
\end{table}

\subsection{The GPU Beam \texorpdfstring{$k$}{k}-opt Search}
\label{sec:appendix_beam}
The GPU beam $k$-opt search is the local search that follows decoding in \ourmethod{} (\cref{sec:method_inference}). A $k$-opt move removes $k$ edges of a route and adds $k$ new edges that give a shorter route. For example, 2-opt, the case $k=2$, replaces a pair of crossing edges with a non-crossing pair. A move with a larger $k$ is stronger, but the number of candidate moves is $\Theta(N^k)$, so a search must find promising moves rather than enumerate all of them.

\noindent\textbf{The \texorpdfstring{$\boldsymbol{k}$}{k}-opt Search Tree of LKH}
LKH \citep{helsgaun2000effective}, which implements the Lin-Kernighan search \citep{lin1973}, finds such moves by building them in steps. As in~\cref{sec:method_inference}, we view this process as a tree whose root is the current route. A step moves one level down the tree. It removes one route edge, which exposes one node of that edge, and adds a new edge from the exposed node to another node, which becomes the ``front node''. The first step can remove any route edge $(n_i,n_j)$ and expose either node, say $n_i$, and we call the other node, $n_j$, the start node. Every later step removes one of the two route edges of the ``front node'' and exposes the other node of that edge. A branch of depth $\ell$ is a sequence of $\ell$ steps from the root, and its gain is $g_\ell = \sum_{s \leq \ell} ( d^{-}_s - d^{+}_s )$, where $d^{-}_s$ and $d^{+}_s$ are the lengths of the edges removed and added at step $s$.

A branch alone does not give a valid route, because its start node has one edge and its ``front node'' has three. Therefore, a step can also close the branch, by joining the exposed node to the start node instead of another node. Every node then has two edges again, and the $\ell$ removals and $\ell$ additions of the closed branch form one $\ell$-opt move, whose gain $g_\ell$ is the decrease in route length. The move is valid when the reconnected route is a single cycle. Otherwise, the closing step is discarded, but the branch stays in the tree, because a deeper closing step can still be valid. A branch can close at every depth, so a $k$-opt search, whose tree has at most $k$ levels, can make an $\ell$-opt move for every $2 \leq \ell \leq k$.

However, some improving moves are out of reach for every branch. The standard example is the double bridge, a 4-opt move made of two 2-opt moves. The first is a \textit{splitting move}, that is, a 2-opt move with the reconnection that reverses no segment and therefore splits the route into two loops. The second is the \textit{fix step}, a 2-opt move that removes one edge from each loop and joins the two loops into a single cycle. Applied alone, the fix step is a splitting move too, so neither of the two is valid alone and the tree discards each of them as a closing step. In addition, no branch contains both, because every step of a branch removes an edge of the ``front node'' of the step before it, and the two moves share no node.

\noindent\textbf{The Sequential Search of LKH}
LKH searches this tree depth-first, that is, it follows one branch as deep as it can, backtracks when the gain is no longer positive, and tries the next branch. When a closed branch gives a shorter route, LKH applies this one move and starts again from the new route. When no branch gives a shorter route, LKH runs a non-sequential search for the moves that no branch contains. The search continues as long as it finds a better route. Each decision therefore depends on the result of the decision before it, and the work differs from route to route. In contrast, a GPU is efficient when many routes perform the same computation at the same moment, so this search runs poorly on a GPU.

\begin{figure}[t]
\centering
    \includegraphics[width=\linewidth]{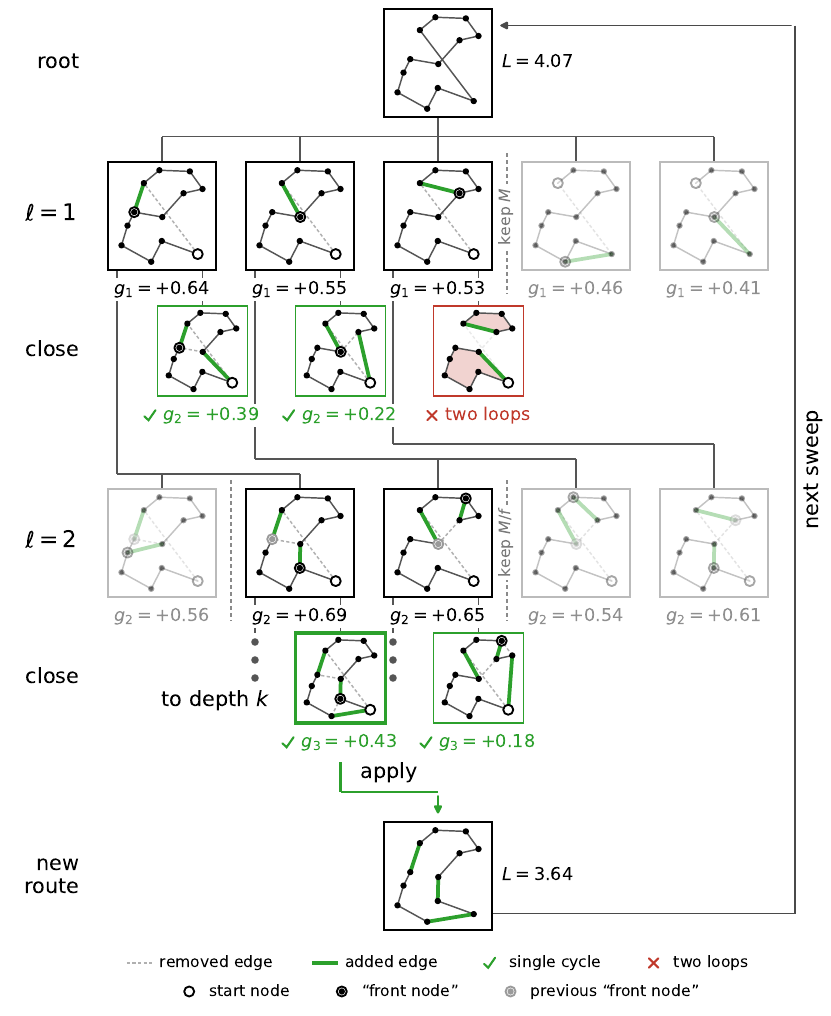}
    \caption{An example sweep of the GPU beam $k$-opt search on a 12-node instance. Every panel draws the same nodes and differs only in its edges. Each level extends every branch of the beam and keeps the branches with the largest gains, and the beam narrows with depth. Each level also closes every branch back to the start node, and a closing that splits the route into two loops is discarded. The sweep applies the kept closures that do not overlap on the route, here only the one with the largest gain. Sweeps repeat until no route improves.}
    \label{fig:beam_tree}
\end{figure}

\noindent\textbf{Our Parallel Beam Search}
Our approach keeps the steps, the gain and the closing rule of LKH. One \textit{sweep} searches the tree once, from the current route down to the largest depth $k$, and then updates the route, and sweeps repeat until no route improves. To fit a GPU, a sweep searches the tree with five changes. \Cref{fig:beam_tree} draws one sweep.

\noindent\textbf{(1) A candidate set for every added edge.} Instead of trying every node in the graph as the new ``front node'', each step adds an edge only to one of the few nodes nearest the exposed node, its \textbf{candidate set}, which loses little because a short route rarely uses a long edge. At the root, this reduces the $\Theta(N^2)$ branches of~\cref{sec:method_inference} to $\Theta(N)$, because the candidate set size is fixed.

\noindent\textbf{(2) A beam of $M$ branches.} Instead of one branch that deepens and backtracks, \textbf{the beam} advances $M$ branches together, level by level, where level $\ell$ holds the branches of depth $\ell$. Each level extends every surviving branch by both route edges of its ``front node'' and by every candidate of the exposed node, computes all gains in one pass, and keeps the $M$ branches with the largest positive gains with one sort. Every route in the batch therefore performs the same computation at the same moment. In addition, the beam narrows with depth by a width-shrink factor $f$, keeping $M$ branches at the first level, $M/f$ at the second, and $M/f^2$ at every deeper level, so deep levels cost less.

\noindent\textbf{(3) All closing steps of a level at once.} Instead of testing one closing step at a time, each level also closes every branch in the beam. Closing a branch only records a move, so the route does not change during a sweep, and the branch itself is still extended to the next level as in (2). The valid closed branches are the \textit{closures} of the level, and across its levels a sweep keeps the closures with the largest gains, up to a fixed number per sweep.

\noindent\textbf{(4) Double-bridge moves in every sweep.} Instead of a separate non-sequential search, every sweep also searches for double-bridge moves. It scores, in one pass, every splitting move that adds an edge between a node and one of the candidates of that node, and keeps the splitting moves with the largest gains. Two of them whose cuts alternate along the route form a double bridge, because the second is the fix step of the first, and the gain of the double bridge is the sum of their gains. A sweep keeps as many double-bridge moves as closures.

\noindent\textbf{(5) Many moves per sweep.} Instead of applying one move at a time, a sweep applies many kept moves together, closures and double-bridge moves alike, as long as they do not overlap on the route. Specifically, each kept move spans a range of route positions, from the first to the last node it touches, and moves whose ranges are at least one position apart can be applied together, because their gains add up and the route stays a single cycle. We find the set of such ranges with the largest total gain by dynamic programming over the ranges sorted by their right end (weighted interval scheduling).

\noindent\textbf{CVRP Variant}
\label{sec:appendix_cvrp_beam}
A CVRP solution is written as one closed route that returns to the depot once per vehicle, so the GPU beam $k$-opt search carries over with two changes. First, the gain of a move becomes its decrease in \textbf{the penalized objective} $c(x) = \tfrac{1}{2}\, d^{\top} x + \rho \sum_{r} \max\big( 0, \, \mathrm{load}(r) - Q \big)$, where $\mathrm{load}(r)$ sums the demands of sub-route $r$, $Q$ is the vehicle capacity and $\rho$ weighs overload, so moves are scored on load as well as on length, and the beam $k$-opt search can pass through a briefly overloaded route to reach a better feasible one. Second, following LKH-3 \citep{helsgaun2017extension}, which extends Lin-Kernighan to capacitated routing, the moves also transfer customers and segments between sub-routes rather than only rearranging edges within one. In addition, a beam sweep is expensive, so instead of running the beam $k$-opt search on each of the $S$ parallel routes, we rank them with a short 2-opt pass and run the search once, on the best of them.

\noindent\textbf{Beam \texorpdfstring{$\boldsymbol{k}$}{k}-opt Search Hyperparameters}
\Cref{tab:appendix_beam_cfg} lists the settings of the GPU beam $k$-opt search, which are the same at every test size. Its last row counts the steps of the 2-opt ranking pass of the CVRP variant, and TSP runs no 2-opt before the beam $k$-opt search.

\begin{table}[ht]
\caption{GPU beam $k$-opt search parameters.}
\label{tab:appendix_beam_cfg}
\centering
\small
\begin{tabular}{l cc}
\toprule
\textbf{Parameter} & \textbf{TSP} & \textbf{CVRP} \\
\midrule
Candidate set size & 15 & 20 \\
Largest $k$ of a $k$-opt move (tree depth) & 5 & 5 \\
Beam width $M$ & 131072 & 32768 \\
Beam width-shrink factor $f$ & 4 & 4 \\
Closures kept per sweep & 256 & 128 \\
2-opt steps before the beam $k$-opt search & 0 & 500 \\
\bottomrule
\end{tabular}
\end{table}

\subsection{Hardware}
\label{sec:appendix_hardware}
All evaluations were conducted on a single NVIDIA L40S GPU, on a machine with an Intel Xeon Gold 6430 CPU, and so was every training run up to target size 1000. The runs that target size 10000 train on a single NVIDIA B200 or H200 GPU. The classical solvers run single-threaded on one core of this CPU.

\section{Configurations of MaskCO and GenSCO}
\label{sec:appendix_baselines}

\noindent\textbf{Correction-Cycle Baselines}
MaskCO \citep{chen2026maskco} and GenSCO \citep{li2025gensco} run their released per-size decoding settings, except the values in~\cref{tab:appendix_cc_baselines}. MaskCO releases no setting above test size 1000, so the size-10000 rows extend its size-1000 settings.

\begin{table}[!ht]
\caption{Correction-cycle baseline settings. Values in parentheses are test sizes.}
\label{tab:appendix_cc_baselines}
\centering
\small
\begin{tabular}{l cc}
\toprule
\textbf{Parameter} & \textbf{TSP} & \textbf{CVRP} \\
\midrule
Correction cycles $C$ & 1, 10 & 1, 10 \\
2-opt moves per cycle & 500 (500), 1000 (1000, 10000) & 6000 \\
\midrule
\multicolumn{3}{l}{\textit{Settings at test size 10000}} \\
Top-k edge slice & $2 \times 10^6$ & $10^6$ \\
Batch size & 1 & 1 \\
\bottomrule
\end{tabular}
\end{table}

\noindent\textbf{MaskCO Self-Train}
MaskCO's optimal-solution-free training \citep{chen2026maskco} replaces the labels of the target size with the model's own solutions. We run it from the released MaskCO model of the source size. In each round, the best model so far by gap solves a set of newly generated target-size instances, and the model is fine-tuned on these solutions with the supervised objective. The gap flattens within four rounds, and the curves of~\cref{fig:curves_cvrp,fig:curves_tsp} use the best model by gap of those rounds.

\section{Additional Results and Ablations}
\label{sec:appendix_results}

\subsection{The Cost of Labels Against Label-Free GRPO}
\label{sec:appendix_label_cost}
\begin{figure}[H]
    \centering
    \includegraphics[width=0.8\linewidth]{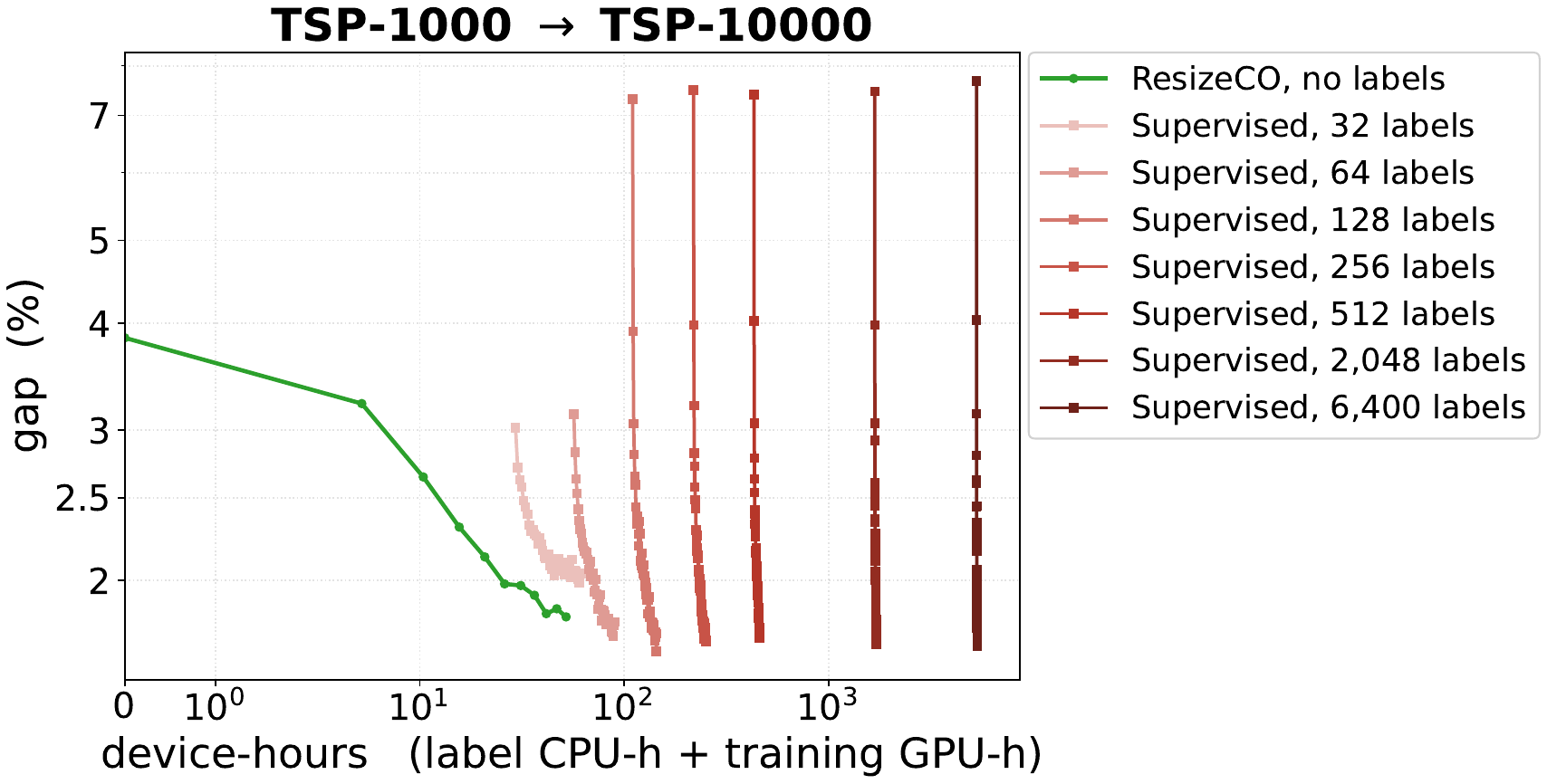}
    \caption{Gap on TSP-10000 against total device hours, the CPU hours of computing the LKH-3 labels plus the GPU hours of training. Every run starts from the same supervised model trained at size 1000. The green line is the label-free GRPO training of \ourmethod{}, and each red line is supervised training with a fixed number of labels.}
    \label{fig:label_cost_tsp}
\end{figure}

\begin{figure}[H]
    \centering
    \includegraphics[width=0.8\linewidth]{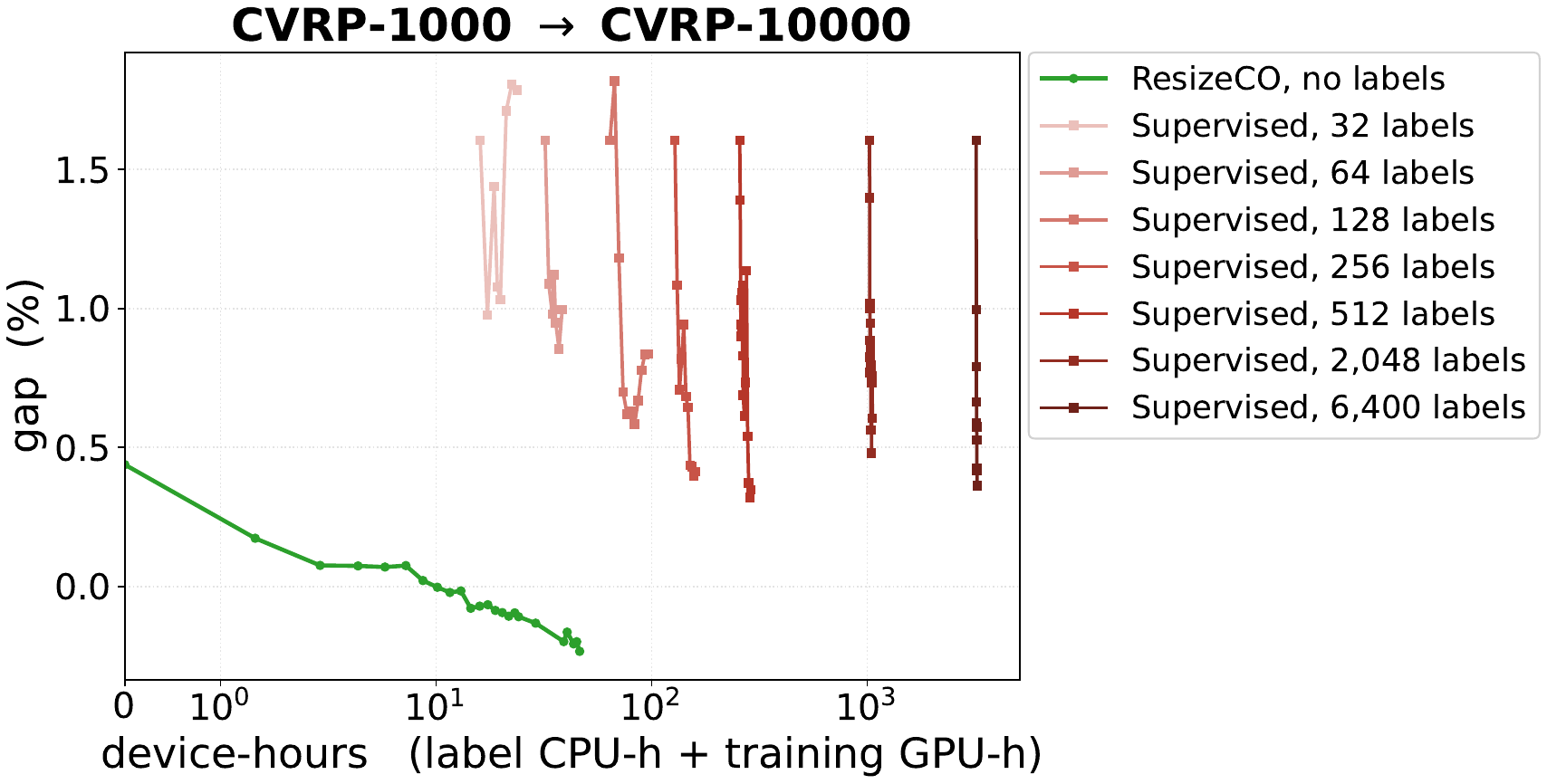}
    \caption{Gap on CVRP-10000 against total device hours, the CPU hours of computing the HGS labels plus the GPU hours of training. Every run starts from the same supervised model trained at size 1000. The green line is the label-free GRPO training of \ourmethod{}, and each red line is supervised training with a fixed number of labels.}
    \label{fig:label_cost_cvrp}
\end{figure}

We compare two ways to train a model for a large target size, both starting from the same supervised model trained at size 1000. The first continues supervised training at the target size, which requires labels computed by a classical solver, LKH-3 for TSP and HGS for CVRP. The second is the label-free GRPO training of \ourmethod{}, which requires no labels. We measure the cost of each in total device hours, the CPU hours of computing the labels plus the GPU hours of training, and report the gap on TSP-10000 and CVRP-10000 as a function of the total device hours in~\cref{fig:label_cost_tsp,fig:label_cost_cvrp}.

\subsection{All Source-to-Target Pairs}
\label{sec:appendix_all_pairs}

This subsection reports all six source-to-target pairs of each problem, grouped by source size. The baseline rows follow~\cref{tab:cvrp_main,tab:tsp_xsize}. Every \ourmethod{} row runs one correction cycle ($C=1$) with one unmasking step ($K=1$), and the rows vary the number of parallel samples from $S=1$ to $S=32$, so each column group shows the gap and the run time at six compute budgets.

\Cref{tab:appendix_cvrp_src100,tab:appendix_tsp_src100} hold the three pairs that start from the models trained at size 100, with two blocks of \ourmethod{} rows. The first block trains one RL run for every pair, as in the main tables. The second block reports a single RL run that starts from the size-100 supervised model and draws the instance size of each training batch from a mix of sizes, so one model serves all three target sizes.

\Cref{tab:appendix_cvrp_src_large,tab:appendix_tsp_src_large} hold the three pairs that start from the models trained at sizes 500 and 1000. They carry only the block with one RL run for every pair, because the mixed-size run starts from the size-100 supervised model.

At target size 10000 the \ourmethod{} models of the first block reach the target in two RL stages. A model of size 100 or 500 is first trained at size 1000 and only then at size 10000, a curriculum that keeps each stage within one order of magnitude of its starting size, while the pair from size 1000 is a single stage. It applies to these rows alone.

\begin{table}[H]
\caption{Size-generalization CVRP results from the models trained at size 100. Conventions follow~\cref{tab:cvrp_main}.}
\label{tab:appendix_cvrp_src100}
\centering
\scriptsize
\setlength{\tabcolsep}{4pt}
\resizebox{\linewidth}{!}{%
\begin{tabular}{l|ccc|ccc|ccc}
\toprule
& \multicolumn{3}{c|}{\textbf{CVRP-100 $\rightarrow$ CVRP-500}} & \multicolumn{3}{c|}{\textbf{CVRP-100 $\rightarrow$ CVRP-1000}} & \multicolumn{3}{c}{\textbf{CVRP-100 $\rightarrow$ CVRP-10000}} \\
\cmidrule(lr){2-4} \cmidrule(lr){5-7} \cmidrule(lr){8-10}
\textbf{Method} & \textbf{Obj} & \textbf{Gap} & \textbf{Time} & \textbf{Obj} & \textbf{Gap} & \textbf{Time} & \textbf{Obj} & \textbf{Gap} & \textbf{Time} \\
\midrule
\multicolumn{10}{l}{\textit{Classical solvers}} \\
HGS (reference) \citep{vidal2022hybrid} & 62.15 & 0.000\% & 55 h & 121.07 & 0.000\% & 197 h & 1120.23 & 0.000\% & 512 h \\
HGS \citep{vidal2022hybrid} & 63.99 & 2.96\% & 52 s & 124.02 & 2.44\% & 2.4 m & 1129.62 & 0.846\% & 2.2 h \\
\midrule
\multicolumn{10}{l}{\textit{Supervised learning}} \\
BQ-NCO (G) \citep{drakulic2023bqnco} & 64.39 & 3.67\% & 26 s & 125.19 & 3.53\% & 36 s & 1221.13 & 9.12\% & 1.3 h \\
LEHD (greedy) \citep{luo2023neural} & 64.40 & 3.63\% & 18 s & 125.41 & 3.66\% & 55 s & \multicolumn{3}{c}{Time Limit} \\
\midrule
\multicolumn{10}{l}{\textit{Label-free learning}} \\
INViT-3V \citep{fang2024invit} & 67.69 & 9.03\% & 6.1 m & 131.78 & 8.97\% & 10.8 m & 1192.43 & 6.49\% & 2.1 h \\
GLOP-G (LKH-3) \citep{ye2024glop} & - & - & - & - & 8.83\%* & 1.4 m* & - & 8.27\%* & 23.9 m* \\
SIT (PRC10) \citep{luo2025boosting} & - & - & - & - & 4.52\%* & 45 s* & - & 4.43\%* & 14.5 m* \\
ReLD (augx8) \citep{huang2025rethinking} & 64.16 & 3.29\% & 21 s & 125.06 & 3.38\% & 40 s & \multicolumn{3}{c}{OOM} \\
\midrule
\multicolumn{10}{l}{\textit{Diffusion-based}} \\
MaskCO (T=128) \citep{chen2026maskco} & 65.79 & 5.90\% & 11 s & 128.25 & 6.03\% & 11 s & 1167.39 & 4.31\% & 1 h \\
MaskCO (T=1280) \citep{chen2026maskco} & 65.39 & 5.25\% & 1.6 m & 127.67 & 5.54\% & 1.6 m & 1163.54 & 3.94\% & 10.2 h \\
\midrule
\multicolumn{10}{l}{\textit{Ours, one RL run per target}} \\
\ourmethod{} (S=1) & 64.10 & 3.18\% & 14 s & 125.13 & 3.43\% & 16 s & 1119.27 & $-$0.056\% & 49.7 m \\
\ourmethod{} (S=2) & 64.02 & 3.04\% & 12 s & 124.89 & 3.23\% & 20 s & 1119.27 & $-$0.057\% & 54.7 m \\
\ourmethod{} (S=4) & 63.91 & 2.88\% & 14 s & 124.86 & 3.20\% & 25 s & 1119.09 & $-$0.073\% & 1.2 h \\
\ourmethod{} (S=8) & 63.88 & 2.83\% & 15 s & 124.76 & 3.12\% & 28 s & 1118.98 & $-$0.082\% & 1.5 h \\
\ourmethod{} (S=16) & 63.82 & 2.72\% & 22 s & 124.64 & 3.02\% & 48 s & 1118.93 & $-$0.088\% & 2.1 h \\
\ourmethod{} (S=32) & 63.81 & 2.71\% & 28 s & 124.63 & 3.01\% & 1.3 m & 1118.78 & $-$0.102\% & 4.1 h \\
\midrule
\multicolumn{10}{l}{\textit{Ours, one RL run for all targets}} \\
\ourmethod{} (S=1) & 64.27 & 3.47\% & 12 s & 124.56 & 2.95\% & 17 s & 1121.31 & 0.130\% & 45.5 m \\
\ourmethod{} (S=2) & 64.22 & 3.38\% & 11 s & 124.49 & 2.90\% & 15 s & 1121.08 & 0.109\% & 50.1 m \\
\ourmethod{} (S=4) & 64.09 & 3.18\% & 13 s & 124.37 & 2.80\% & 17 s & 1120.90 & 0.092\% & 58.7 m \\
\ourmethod{} (S=8) & 64.04 & 3.10\% & 14 s & 124.29 & 2.73\% & 23 s & 1120.82 & 0.085\% & 1.3 h \\
\ourmethod{} (S=16) & 63.98 & 2.99\% & 17 s & 124.23 & 2.68\% & 37 s & 1120.70 & 0.074\% & 2 h \\
\ourmethod{} (S=32) & 63.96 & 2.96\% & 19 s & 124.17 & 2.63\% & 1.1 m & 1120.62 & 0.067\% & 3.6 h \\
\bottomrule
\end{tabular}}
\end{table}

\begin{table}[H]
\caption{Size-generalization TSP results from the models trained at size 100. Conventions follow~\cref{tab:tsp_xsize}.}
\label{tab:appendix_tsp_src100}
\centering
\scriptsize
\setlength{\tabcolsep}{4pt}
\resizebox{\linewidth}{!}{%
\begin{tabular}{l|ccc|ccc|ccc}
\toprule
& \multicolumn{3}{c|}{\textbf{TSP-100 $\rightarrow$ TSP-500}} & \multicolumn{3}{c|}{\textbf{TSP-100 $\rightarrow$ TSP-1000}} & \multicolumn{3}{c}{\textbf{TSP-100 $\rightarrow$ TSP-10000}} \\
\cmidrule(lr){2-4} \cmidrule(lr){5-7} \cmidrule(lr){8-10}
\textbf{Method} & \textbf{Obj} & \textbf{Gap} & \textbf{Time} & \textbf{Obj} & \textbf{Gap} & \textbf{Time} & \textbf{Obj} & \textbf{Gap} & \textbf{Time} \\
\midrule
\multicolumn{10}{l}{\textit{Classical solvers}} \\
Concorde / LKH-3 (reference) & 16.55 & 0.000\% & 37.66 m & 23.12 & 0.000\% & 6.65 h & 71.83 & 0.000\% & 8.8 h \\
LKH-3 (trials=32) \citep{helsgaun2017extension} & 16.55 & 0.026\% & 22 s & 23.13 & 0.036\% & 56 s & 71.81 & $-$0.023\% & 7.7 m \\
LKH-3 (trials=1) \citep{helsgaun2017extension} & 16.57 & 0.118\% & 8.6 s & 23.15 & 0.139\% & 23 s & 71.89 & 0.085\% & 5.6 m \\
\midrule
\multicolumn{10}{l}{\textit{Supervised learning}} \\
BQ-NCO (G) \citep{drakulic2023bqnco} & 16.74 & 1.19\% & 23 s & 23.65 & 2.29\% & 54 s & 89.07 & 24.00\% & 9.4 m \\
LEHD (greedy) \citep{luo2023neural} & 16.80 & 1.53\% & 15 s & 23.84 & 3.12\% & 1.7 m & 91.47 & 27.34\% & 2.8 h \\
\midrule
\multicolumn{10}{l}{\textit{Label-free learning}} \\
INViT-3V \citep{fang2024invit} & 17.27 & 4.38\% & 4.9 m & 24.30 & 5.10\% & 9.8 m & 76.57 & 6.61\% & 33.2 m \\
GLOP \citep{ye2024glop} & 17.07 & 3.20\% & 19 s & 24.01 & 3.85\% & 38 s & 75.61 & 5.27\% & 29 s \\
SIT (PRC10) \citep{luo2025boosting} & 16.70 & 0.947\% & 28 s & 23.39 & 1.18\% & 1.3 m & 74.01 & 3.04\% & 2.1 m \\
\midrule
\multicolumn{10}{l}{\textit{Diffusion-based}} \\
DIFUSCO (T$_s$=50, G) \citep{sun2023difusco} & 17.09 & 3.27\% & 2.2 m & 24.09 & 4.19\% & 10 m & 76.68 & 6.75\% & 41.7 m \\
DIFUSCO (T$_s$=100, S=4) \citep{sun2023difusco} & 16.97 & 2.53\% & 21.6 m & 23.94 & 3.53\% & 1.2 h & \multicolumn{3}{c}{Time Limit} \\
Fast T2T (T$_s$=5, T$_g$=5, G) \citep{li2024fastt2t} & 17.30 & 4.58\% & 2.5 m & 24.39 & 5.48\% & 10.9 m & \multicolumn{3}{c}{Time Limit} \\
Fast T2T (T$_s$=5, T$_g$=5, S=4) \citep{li2024fastt2t} & 17.16 & 3.71\% & 11.2 m & 24.22 & 4.75\% & 24.6 m & \multicolumn{3}{c}{OOM} \\
GenSCO (C=1, S=8) \citep{li2025gensco} & 17.19 & 3.88\% & 2.5 s & 24.96 & 7.99\% & 20 s & 351.17 & 388.91\% & 2.8 m \\
GenSCO (C=10, S=8) \citep{li2025gensco} & 17.05 & 3.06\% & 25 s & 24.80 & 7.26\% & 3.3 m & 350.60 & 388.11\% & 28.1 m \\
MaskCO (T=8/16) \citep{chen2026maskco} & 17.21 & 3.99\% & 1 s & 24.99 & 8.10\% & 28 s & 332.06 & 362.30\% & 3.1 m \\
MaskCO (T=80/160) \citep{chen2026maskco} & 17.07 & 3.18\% & 10 s & 24.80 & 7.27\% & 3.9 m & 321.28 & 347.30\% & 31 m \\
\midrule
\multicolumn{10}{l}{\textit{Ours, one RL run per target}} \\
\ourmethod{} (S=1) & 16.60 & 0.329\% & 2.6 s & 23.31 & 0.813\% & 7.5 s & 74.72 & 4.03\% & 43 s \\
\ourmethod{} (S=2) & 16.59 & 0.261\% & 3.3 s & 23.28 & 0.689\% & 9.6 s & 74.62 & 3.89\% & 1.3 m \\
\ourmethod{} (S=4) & 16.58 & 0.213\% & 4.8 s & 23.26 & 0.599\% & 16 s & 74.57 & 3.82\% & 2.2 m \\
\ourmethod{} (S=8) & 16.58 & 0.180\% & 12 s & 23.24 & 0.541\% & 30 s & 74.55 & 3.80\% & 4 m \\
\ourmethod{} (S=16) & 16.57 & 0.152\% & 17 s & 23.24 & 0.515\% & 52 s & 74.49 & 3.71\% & 7.7 m \\
\ourmethod{} (S=32) & 16.57 & 0.144\% & 38 s & 23.23 & 0.482\% & 1.9 m & 74.45 & 3.65\% & 15.3 m \\
\midrule
\multicolumn{10}{l}{\textit{Ours, one RL run for all targets}} \\
\ourmethod{} (S=1) & 16.61 & 0.360\% & 2.2 s & 23.34 & 0.968\% & 8.1 s & 74.90 & 4.27\% & 48 s \\
\ourmethod{} (S=2) & 16.59 & 0.283\% & 4.1 s & 23.31 & 0.838\% & 11 s & 74.80 & 4.14\% & 1.4 m \\
\ourmethod{} (S=4) & 16.58 & 0.214\% & 5.9 s & 23.29 & 0.746\% & 18 s & 74.76 & 4.09\% & 2.6 m \\
\ourmethod{} (S=8) & 16.57 & 0.174\% & 13 s & 23.28 & 0.690\% & 33 s & 74.74 & 4.06\% & 4.7 m \\
\ourmethod{} (S=16) & 16.57 & 0.147\% & 23 s & 23.27 & 0.645\% & 59 s & 74.71 & 4.01\% & 9 m \\
\ourmethod{} (S=32) & 16.57 & 0.138\% & 43 s & 23.26 & 0.615\% & 1.9 m & 74.67 & 3.95\% & 17 m \\
\bottomrule
\end{tabular}}
\end{table}

\begin{table}[H]
\caption{Size-generalization CVRP results from the models trained at sizes 500 and 1000. Conventions follow~\cref{tab:cvrp_main}.}
\label{tab:appendix_cvrp_src_large}
\centering
\scriptsize
\setlength{\tabcolsep}{4pt}
\resizebox{\linewidth}{!}{%
\begin{tabular}{l|ccc|ccc|ccc}
\toprule
& \multicolumn{3}{c|}{\textbf{CVRP-500 $\rightarrow$ CVRP-1000}} & \multicolumn{3}{c|}{\textbf{CVRP-1000 $\rightarrow$ CVRP-10000}} & \multicolumn{3}{c}{\textbf{CVRP-500 $\rightarrow$ CVRP-10000}} \\
\cmidrule(lr){2-4} \cmidrule(lr){5-7} \cmidrule(lr){8-10}
\textbf{Method} & \textbf{Obj} & \textbf{Gap} & \textbf{Time} & \textbf{Obj} & \textbf{Gap} & \textbf{Time} & \textbf{Obj} & \textbf{Gap} & \textbf{Time} \\
\midrule
\multicolumn{10}{l}{\textit{Classical solvers}} \\
HGS (reference) \citep{vidal2022hybrid} & 121.07 & 0.000\% & 197 h & 1120.23 & 0.000\% & 512 h & 1120.23 & 0.000\% & 512 h \\
HGS \citep{vidal2022hybrid} & 124.02 & 2.44\% & 2.4 m & 1129.62 & 0.846\% & 2.2 h & 1129.62 & 0.846\% & 2.2 h \\
\midrule
\multicolumn{10}{l}{\textit{Supervised learning}} \\
BQ-NCO (G) \citep{drakulic2023bqnco} & - & - & - & - & - & - & - & - & - \\
LEHD (greedy) \citep{luo2023neural} & - & - & - & - & - & - & - & - & - \\
\midrule
\multicolumn{10}{l}{\textit{Label-free learning}} \\
INViT-3V \citep{fang2024invit} & 131.78 & 8.97\% & 10.8 m & 1192.43 & 6.49\% & 2.1 h & 1192.43 & 6.49\% & 2.1 h \\
GLOP-G (LKH-3) \citep{ye2024glop} & - & 8.83\%* & 1.4 m* & - & 8.27\%* & 23.9 m* & - & 8.27\%* & 23.9 m* \\
SIT (PRC10) \citep{luo2025boosting} & - & 4.52\%* & 45 s* & - & 4.43\%* & 14.5 m* & - & 4.43\%* & 14.5 m* \\
ReLD (augx8) \citep{huang2025rethinking} & 125.06 & 3.38\% & 40 s & \multicolumn{3}{c|}{OOM} & \multicolumn{3}{c}{OOM} \\
\midrule
\multicolumn{10}{l}{\textit{Diffusion-based}} \\
MaskCO (T=128) \citep{chen2026maskco} & 125.79 & 3.98\% & 11 s & 1197.55 & 7.97\% & 1 h & 1639.21 & 45.15\% & 1 h \\
MaskCO (T=1280) \citep{chen2026maskco} & 125.79 & 3.98\% & 1.6 m & 1146.56 & 2.36\% & 10.6 h & 1127.41 & 0.686\% & 10.5 h \\
\midrule
\multicolumn{10}{l}{\textit{Ours, one RL run per target}} \\
\ourmethod{} (S=1) & 124.35 & 2.76\% & 13 s & 1118.83 & $-$0.096\% & 49.9 m & 1117.51 & $-$0.217\% & 46 m \\
\ourmethod{} (S=2) & 124.21 & 2.64\% & 14 s & 1118.60 & $-$0.117\% & 55.4 m & 1117.36 & $-$0.231\% & 50.9 m \\
\ourmethod{} (S=4) & 124.08 & 2.54\% & 18 s & 1118.50 & $-$0.125\% & 1 h & 1117.27 & $-$0.240\% & 1 h \\
\ourmethod{} (S=8) & 124.00 & 2.48\% & 26 s & 1118.32 & $-$0.143\% & 1.4 h & 1117.25 & $-$0.241\% & 1.3 h \\
\ourmethod{} (S=16) & 123.96 & 2.44\% & 44 s & 1118.20 & $-$0.151\% & 2.2 h & 1117.09 & $-$0.256\% & 1.9 h \\
\ourmethod{} (S=32) & 123.82 & 2.32\% & 1.2 m & 1118.08 & $-$0.162\% & 4.3 h & 1117.06 & $-$0.259\% & 3.8 h \\
\bottomrule
\end{tabular}}
\end{table}

\begin{table}[H]
\caption{Size-generalization TSP results from the models trained at sizes 500 and 1000. Conventions follow~\cref{tab:tsp_xsize}.}
\label{tab:appendix_tsp_src_large}
\centering
\scriptsize
\setlength{\tabcolsep}{4pt}
\resizebox{\linewidth}{!}{%
\begin{tabular}{l|ccc|ccc|ccc}
\toprule
& \multicolumn{3}{c|}{\textbf{TSP-500 $\rightarrow$ TSP-1000}} & \multicolumn{3}{c|}{\textbf{TSP-1000 $\rightarrow$ TSP-10000}} & \multicolumn{3}{c}{\textbf{TSP-500 $\rightarrow$ TSP-10000}} \\
\cmidrule(lr){2-4} \cmidrule(lr){5-7} \cmidrule(lr){8-10}
\textbf{Method} & \textbf{Obj} & \textbf{Gap} & \textbf{Time} & \textbf{Obj} & \textbf{Gap} & \textbf{Time} & \textbf{Obj} & \textbf{Gap} & \textbf{Time} \\
\midrule
\multicolumn{10}{l}{\textit{Classical solvers}} \\
Concorde / LKH-3 (reference) & 23.12 & 0.000\% & 6.65 h & 71.83 & 0.000\% & 8.8 h & 71.83 & 0.000\% & 8.8 h \\
LKH-3 (trials=32) \citep{helsgaun2017extension} & 23.13 & 0.036\% & 56 s & 71.81 & $-$0.023\% & 7.7 m & 71.81 & $-$0.023\% & 7.7 m \\
LKH-3 (trials=1) \citep{helsgaun2017extension} & 23.15 & 0.139\% & 23 s & 71.89 & 0.085\% & 5.6 m & 71.89 & 0.085\% & 5.6 m \\
\midrule
\multicolumn{10}{l}{\textit{Supervised learning}} \\
BQ-NCO (G) \citep{drakulic2023bqnco} & - & - & - & - & - & - & - & - & - \\
LEHD (greedy) \citep{luo2023neural} & - & - & - & - & - & - & - & - & - \\
\midrule
\multicolumn{10}{l}{\textit{Label-free learning}} \\
INViT-3V \citep{fang2024invit} & 24.30 & 5.10\% & 9.8 m & 76.57 & 6.61\% & 33.2 m & 76.57 & 6.61\% & 33.2 m \\
GLOP \citep{ye2024glop} & 24.01 & 3.85\% & 38 s & 75.61 & 5.27\% & 29 s & 75.61 & 5.27\% & 29 s \\
SIT (PRC10) \citep{luo2025boosting} & 23.39 & 1.18\% & 1.3 m & 74.01 & 3.04\% & 2.1 m & 74.01 & 3.04\% & 2.1 m \\
\midrule
\multicolumn{10}{l}{\textit{Diffusion-based}} \\
DIFUSCO (T$_s$=50, G) \citep{sun2023difusco} & 23.64 & 2.27\% & 9.6 m & 74.11 & 3.18\% & 29.6 m & 75.15 & 4.63\% & 33.3 m \\
DIFUSCO (T$_s$=100, S=4) \citep{sun2023difusco} & 23.48 & 1.56\% & 1.2 h & \multicolumn{3}{c|}{Time Limit} & \multicolumn{3}{c}{Time Limit} \\
Fast T2T (T$_s$=5, T$_g$=5, G) \citep{li2024fastt2t} & 23.46 & 1.48\% & 8.9 m & 75.20 & 4.70\% & 3.6 h & 76.16 & 6.03\% & 3.9 h \\
Fast T2T (T$_s$=5, T$_g$=5, S=4) \citep{li2024fastt2t} & 23.37 & 1.10\% & 20.1 m & \multicolumn{3}{c|}{OOM} & \multicolumn{3}{c}{OOM} \\
GenSCO (C=1, S=8) \citep{li2025gensco} & 23.17 & 0.245\% & 20 s & 89.77 & 24.97\% & 3.4 m & 118.85 & 65.46\% & 3.3 m \\
GenSCO (C=10, S=8) \citep{li2025gensco} & 23.15 & 0.138\% & 3.3 m & 89.34 & 24.38\% & 34 m & 118.41 & 64.85\% & 33.3 m \\
MaskCO (T=16) \citep{chen2026maskco} & 23.27 & 0.662\% & 23 s & 75.60 & 5.26\% & 3.4 m & 75.79 & 5.52\% & 3.4 m \\
MaskCO (T=160) \citep{chen2026maskco} & 23.15 & 0.138\% & 3.8 m & 75.24 & 4.75\% & 33.9 m & 75.41 & 4.99\% & 33.8 m \\
\midrule
\multicolumn{10}{l}{\textit{Ours, one RL run per target}} \\
\ourmethod{} (S=1) & 23.17 & 0.208\% & 5.9 s & 73.30 & 2.05\% & 34 s & 74.10 & 3.16\% & 42 s \\
\ourmethod{} (S=2) & 23.16 & 0.165\% & 7.6 s & 73.23 & 1.95\% & 53 s & 74.01 & 3.04\% & 1.1 m \\
\ourmethod{} (S=4) & 23.15 & 0.136\% & 11 s & 73.19 & 1.90\% & 1.4 m & 73.99 & 3.01\% & 1.9 m \\
\ourmethod{} (S=8) & 23.14 & 0.112\% & 17 s & 73.16 & 1.85\% & 2.5 m & 73.95 & 2.96\% & 3.5 m \\
\ourmethod{} (S=16) & 23.14 & 0.099\% & 31 s & 73.14 & 1.83\% & 4.6 m & 73.92 & 2.91\% & 6.6 m \\
\ourmethod{} (S=32) & 23.14 & 0.096\% & 1 m & 73.14 & 1.83\% & 8.8 m & 73.89 & 2.87\% & 12.8 m \\
\bottomrule
\end{tabular}}
\end{table}

\subsection{Component Ablation from TSP-100 to TSP-1000}
\label{sec:appendix_more_pairs}
\Cref{tab:abl_components} ablates the components of \ourmethod{} on two source-to-target pairs. \Cref{tab:appendix_abl_more} repeats the ablation on a third pair, TSP-100 $\rightarrow$ TSP-1000. Each row removes a single component. Removing any component increases the gap, most of all label-free GRPO, from 0.813\% to 3.23\%.

\begin{table}[H]
\caption{Component ablation of \ourmethod{} from TSP-100 to TSP-1000. Conventions follow~\cref{tab:abl_components}.}
\label{tab:appendix_abl_more}
\centering
\footnotesize
\setlength{\tabcolsep}{4pt}
\begin{tabular}{l|cc}
\toprule
& \multicolumn{2}{c}{\textbf{TSP-100 $\rightarrow$ TSP-1000}} \\
\cmidrule(lr){2-3}
\textbf{Method} & \textbf{Gap} & \textbf{Time} \\
\midrule
\textbf{\ourmethod{} (full)} & \textbf{0.813\%} & \textbf{7.5 s} \\
\midrule
w/o label-free GRPO & 3.23\% & 10.7 s \\
w/o small-model initialization & 2.67\% & 10.6 s \\
w/o Graph Transformer encoder & 2.42\% & 8.7 s \\
w/o GPU beam $k$-opt search & 1.29\% & 7.7 s \\
\bottomrule
\end{tabular}
\end{table}

\subsection{Why Does Node-Walk Decoding Help Training?}
\label{sec:method_walk}
\label{sec:appendix_decoder_runs}

\begin{figure}[!tb]
    \centering
    \includegraphics[width=0.6\linewidth]{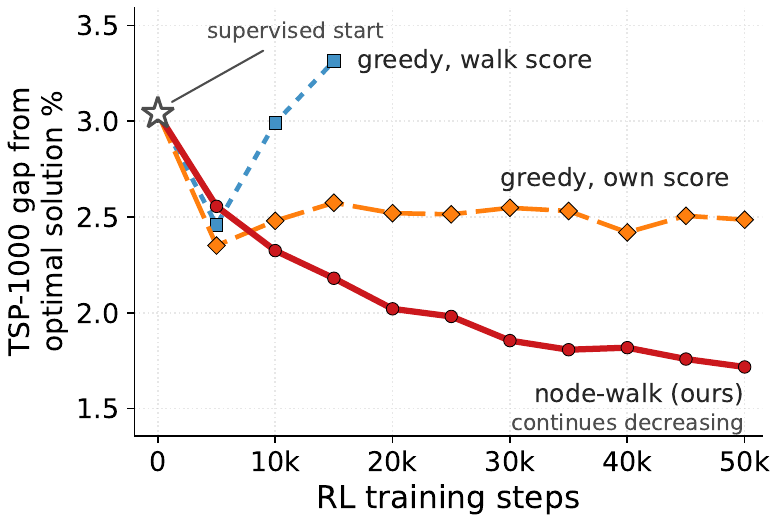}
    \caption{Three GRPO runs from the same supervised model. The runs differ only in how routes are sampled and scored during RL training. \textbf{Blue: }sampling with the greedy decoder and scoring with the node-walk likelihood, the TSP-1000 gap rises above the supervised starting point. \textbf{Orange: }sampling with the greedy decoder and scoring with its own exact likelihood, the gap stays flat. \textbf{Red: }sampling and scoring with node-walk decoding (our approach), the gap falls steadily.}
    \label{fig:walk_vs_insert}
\end{figure}

Previous heatmap-based CO solvers decode with the greedy decoder \citep{sun2023difusco}. The decoder a GRPO run uses for its rollouts decides whether the training improves the model at all, so we compare three runs from the same supervised model that differ only in the decoder (\cref{fig:walk_vs_insert}).
Node-walk decoding scored by its exact likelihood, Eq.~\eqref{eq:walk}, improves steadily. Greedy decoding scored by that same likelihood falls below the supervised model it started from, because the node-walk likelihood never sampled the routes a random edge ranking produces, which biases the gradient. \textbf{Greedy decoding scored by the exact probability of its own ranking removes the bias, and the collapse with it, yet the model still does not learn.} The obstacle is therefore not the validity of the gradient.

We attribute the failure to the way each decoder normalizes the transition probability $p_\theta$. The node-walk decoder weighs an edge against the edges from $n_j$ to the unvisited nodes $U_j$, giving $p^{t}_{\theta,\, n_j n_{j+1}} / \sum_{n' \in U_j} p^{t}_{\theta,\, n_j n'}$, a normalization within a row, as in the supervised loss. In contrast, the greedy decoder weighs it against the set $\mathcal{B}$ of all edges still masked in the graph, giving $p^{t}_{\theta,\, uv} / \sum_{(u',v') \in \mathcal{B}} p^{t}_{\theta,\, u'v'}$, which compares unrelated edges. At inference, the greedy decoder works well, but we deploy the node-walk decoder, so training and inference use the same decoder.

\section{Derivation of the Training Loss}
\label{sec:appendix_loss}

This appendix derives the training loss of~\cref{sec:method_rl}. We start from the standard GRPO loss of Eq.~\eqref{eq:grpo_clip}, written for our $N$ decisions,
\[
    \mathcal{L}_{\mathrm{GRPO}}(\theta) = \frac{1}{G} \sum_{g=1}^{G} \frac{1}{N} \sum_{j=1}^{N} \Big[ - \min\big( r^g_j(\theta)\, A_g , \mathrm{clip}\big(r^g_j, 1{-}\varepsilon, 1{+}\varepsilon\big) A_g \big) + \beta \, D_{\mathrm{KL}}\big( \pi_\theta \,\Vert\, \pi_{\mathrm{ref}} \big)(s_j) \Big] ,
\]
where $r^g_j(\theta) = \pi_\theta(a_j \mid s_j) \,/\, \pi_{\theta_{\mathrm{old}}}(a_j \mid s_j)$ is the ratio between the current policy and the policy that sampled the group. Throughout this appendix, $\pi_\theta(a_j \mid s_j)$ is short for the node-walk policy $\pi_\theta(a_j \mid s_j ; p^{t}_\theta)$ of Eq.~\eqref{eq:walk-step}.

\textbf{In our case, we adapt the loss in two ways.
}
First, we take a \textbf{single update per batch} and resample fresh rollouts, which are cheap to draw. The rollouts are therefore always sampled by the current policy, so $\theta_{\mathrm{old}} = \theta$. The ratios $r^g_j$ then equal one, the clipping never takes effect, and the first term of Eq.~\eqref{eq:grpo_clip} reduces to $-\,r^g_j(\theta)\, A_g$. This term has the same gradient as $-\,A_g \log \pi_\theta(a_j \mid s_j)$, because
\begin{equation}
    \nabla_\theta\, r^g_j(\theta)
    = \frac{\nabla_\theta\, \pi_\theta(a_j \mid s_j)}{\pi_{\theta_{\mathrm{old}}}(a_j \mid s_j)}
    = \frac{\nabla_\theta\, \pi_\theta(a_j \mid s_j)}{\pi_\theta(a_j \mid s_j)}
    = \nabla_\theta \log \pi_\theta(a_j \mid s_j) .
\end{equation}
The first equality treats $\pi_{\theta_{\mathrm{old}}}$ as a constant, because $\theta_{\mathrm{old}}$ is fixed during the update. The second substitutes $\theta_{\mathrm{old}} = \theta$. The third is the derivative of a logarithm. Summing over the $N$ decisions and using the route log-probability of Eq.~\eqref{eq:walk}, the loss of one route becomes its advantage times the route log-probability,
\begin{equation}
    -\,\frac{1}{N}\, A_g \sum_{j=1}^{N} \log \pi_\theta(a_j \mid s_j)
    = -\,\frac{1}{N}\, A_g \log p_\theta(\mathrm{route}_g) .
\end{equation}

Second, we apply the \textbf{KL penalty to the whole edge heatmap}, not only to the visited states. This is because a gradient update changes the model weights $\theta$, and with them every row of the heatmap, even rows that no rollout read. At inference, the node-walk decoder reads such rows as well, so the whole heatmap must stay near the reference $p_{\mathrm{ref}}$. Specifically, the per-state KL of Eq.~\eqref{eq:grpo_clip} is replaced by the row average
\begin{equation}
    D_{\mathrm{KL}}\big( p_\theta \,\Vert\, p_{\mathrm{ref}} \big) = \frac{1}{N} \sum_{i=1}^{N} D_{\mathrm{KL}}\big( p_{\theta,\, i} \,\Vert\, p_{\mathrm{ref},\, i} \big) ,
\end{equation}
where $p_{\theta,\, i}$ is row $i$ of the heatmap, averaged over all heatmaps in the rollout.

Putting the two changes together, \textbf{the final loss} averages the route loss over the group and adds the KL penalty,
\begin{equation}
    \mathcal{L}_{\mathrm{\ourmethod}}(\theta) = - \frac{1}{N} \frac{1}{G} \sum_{g=1}^{G} A_g \log p_\theta(\mathrm{route}_g) + \beta \, D_{\mathrm{KL}}\big( p_\theta \,\Vert\, p_{\mathrm{ref}} \big) .
\end{equation}

\end{document}